\documentclass[11pt]{article}

\usepackage[letterpaper,margin=1in]{geometry}
\usepackage[utf8]{inputenc}
\usepackage[T1]{fontenc}
\usepackage[numbers,sort&compress]{natbib}
\usepackage{url}
\usepackage{booktabs}
\usepackage{amsfonts}
\usepackage{nicefrac}
\usepackage{microtype}
\usepackage{xcolor}
\usepackage{amsmath}
\usepackage{amssymb}
\usepackage{graphicx}
\usepackage{algorithm}
\usepackage{algorithmic}
\usepackage{subcaption}
\usepackage{wrapfig}
\usepackage{tabularx}
\usepackage{array}
\usepackage{makecell}
\usepackage{float}
\usepackage{cleveref}

\title{ReFPO: Reflow Regularization for Flow Matching Policy Gradients}

\newcommand{\corr}{\textsuperscript{\textdagger}}
\newcommand{\ising}{\textsuperscript{1}}
\newcommand{\cuhk}{\textsuperscript{2}}
\newcommand{\fnii}{\textsuperscript{3}}
\newcommand{\affilall}{\textsuperscript{1,2,3}}
\newcommand{\affilthirteen}{\textsuperscript{1,3}}

\renewenvironment{abstract}%
{%
  \vspace{0.075in}%
  \begin{center}%
    {\large\bfseries Abstract}%
  \end{center}%
  \vspace{-0.5em}%
  \begin{quote}%
}
{%
  \end{quote}%
  \vspace{1ex}%
}

\author{
  \begin{tabular}{c}
    Ge Wang\affilall \quad
    Yibo Peng\affilall \quad
    Fan Feng\affilthirteen \quad
    Shenhao Yan\affilthirteen\\
    Chengsi Yao\affilthirteen \quad
    Jiahao Yang\affilthirteen \quad
    Honghao Cai\affilall \quad
    Yiming Zhao\ising \quad
    Xi Li\ising\\
    Jinke Ren\textsuperscript{2,3} \quad
    Shuguang Cui\textsuperscript{2,3} \quad
    Yatong Han\textsuperscript{1,\textdagger} \quad
    Zhen Li\textsuperscript{2,3,\textdagger}\\[0.45em]
    \small \ising\,Ising AI \quad \cuhk\,CUHK-Shenzhen \quad \fnii\,FNii-Shenzhen\\[0.2em]
    \small \corr Corresponding authors.
  \end{tabular}
}
\date{}

\begin{document}

\maketitle

\begin{abstract}
We present Reflow-regularized Flow Matching Policy Gradients (ReFPO), a simple online RL method that adds explicit Reflow regularization to FPO for efficient flow-based control.
 We uncover a key structural property: the gradient updates in Flow Matching Policy Gradients (FPO) can be interpreted as an implicit advantage-weighted Reflow process, providing a new geometric perspective on flow-based policy gradients. Building on this insight, ReFPO introduces an explicit geometric regularizer that can be implemented with a single line of code change without incurring additional computational overhead or auxiliary distillation stages. By synergizing advantage-guided updates with path rectification, our method reduces CFM proxy-ratio spikes, stabilizes PPO-style training, and enables high-fidelity one-step inference that often matches or exceeds multi-step performance. We experimentally demonstrate that ReFPO improves average performance and discretization robustness across GridWorld, MuJoCo Playground, and high-dimensional Humanoid Control tasks, providing a scalable and stable approach for generative policies in complex physical simulations.
\end{abstract}

\section{Introduction}

\begin{wrapfigure}{r}{0.56\textwidth}
  \centering
  \vspace{-0.8em}
  \includegraphics[width=0.54\textwidth]{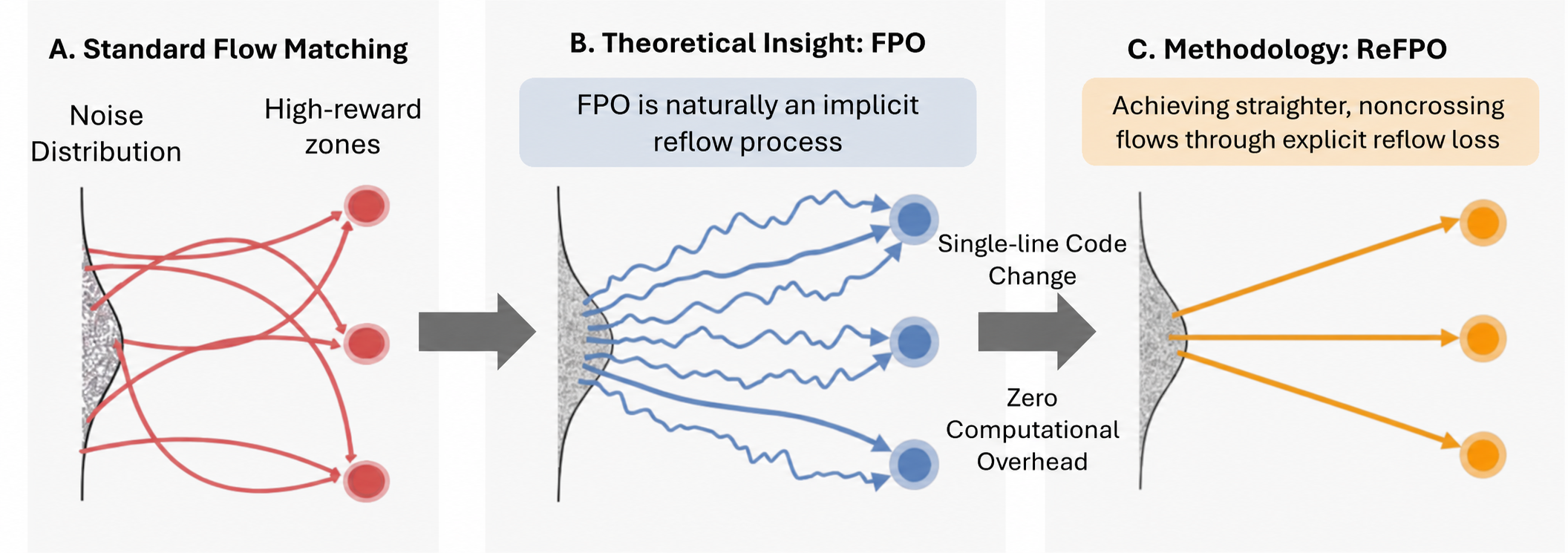}
  \caption{\textbf{Overview of ReFPO.} }
  \label{figure:refpo_overview}
  \vspace{-1em}
\end{wrapfigure}

Generative models, particularly flow matching and diffusion, have become powerful policy representations in reinforcement learning (RL) by enabling the capture of complex, multimodal action distributions. Unlike traditional Gaussian policies, flow-based policies can represent non-convex behaviors essential for high-dimensional tasks. However, their practical deployment is limited by their iterative sampling nature. Generating a single action often requires multiple ODE integration steps, introducing a significant latency barrier for real-time robotic deployment~\cite{chi2025diffusion}.

To mitigate this gap, prior acceleration techniques such as Rectified Flow~\cite{liu2023flow}, Consistency Models~\cite{song2023consistencymodels}, or multi-stage distillation~\cite{yin2024one} have been proposed. While effective in supervised computer vision, directly grafting these methods onto RL often results in cumbersome, multi-stage pipelines that struggle with the non-stationary landscape of online optimization. These ``force-fitted" combinations introduce significant computational overhead and additional training complexity, lacking the elegance required for seamless RL integration.

As illustrated in Figure~\ref{figure:refpo_overview}, we introduce \textbf{Reflow-regularized Flow Matching Policy Gradients (ReFPO)}, a framework built on the observation that the FPO proxy can be viewed as an advantage-weighted use of the Conditional Flow Matching (CFM) loss. In online RL, the action-noise pairs used by this CFM loss are repeatedly generated by the policy itself. Therefore, beyond optimizing the reward-weighted policy objective, FPO also induces a self-iterative Reflow-style straightening effect on its own sampled trajectories. This gives a natural and elegant connection: the same CFM objective that makes FPO a tractable PPO proxy also gives it a self-sampled Reflow structure when placed inside an online RL loop. ReFPO makes this implicit geometry explicit and turns it into a lightweight regularizer for stable and efficient flow-policy learning.

% \begin{figure}
%   \centering
%   \includegraphics[width=\columnwidth]{figures/Figure1b.pdf}
%   \caption{\textbf{Overview of ReFPO.} (A) Standard Flow Matching often results in curved, inefficient probability paths when mapping noise to high-reward zones. (B) We provide a theoretical insight that FPO is naturally an implicit reflow process, though it lacks explicit path straightening. (C) Our proposed ReFPO achieves optimal transport through an explicit reflow loss, enabling straight sampling paths with a single-line code change and zero additional computational overhead.}
%   \label{figure:refpo_overview}
% \end{figure}

However, this implicit Reflow effect is not explicitly controlled. While the advantage-weighted update straightens flows for high-reward actions, it simultaneously distorts the flow field for actions with negative advantages. This imbalance can increase local bending of the fitted velocity field, amplify one-step discretization errors, and make the CFM-based proxy ratio more volatile during PPO-style updates. To resolve this, ReFPO reuses the CFM loss already computed in the FPO pipeline as an explicit, unweighted Reflow regularization term. By incorporating this term into the total objective, we bias the learned flow field toward straighter paths regardless of the advantage sign. This provides a crucial geometric anchor that improves one-step consistency, reduces proxy-ratio spikes, and yields straighter, more efficient flows.

The primary contributions of this paper are threefold. First, we identify a natural and elegant connection between FPO and Reflow: the CFM proxy used by FPO has the same geometric form as a Reflow objective on policy-generated action-noise pairs, giving FPO an online, advantage-weighted Reflow interpretation. More broadly, this perspective suggests that Reflow-style geometry is relevant to online flow-matching policy optimization whenever the policy is trained on self-sampled action-noise pairs with reward- or advantage-dependent reweighting. Second, we introduce explicit Reflow Regularization, which reuses the existing CFM loss to make the learned velocity field straighter. This lightweight regularizer enables deployment with fewer denoising steps, preserves multimodal policy expressivity through the reward-guided policy objective, and reduces the collapse risk associated with uncontrolled self-rectification. Since the regularizer only requires the same action-noise interpolants already used by flow-matching policies, it can serve as a general geometric stabilizer for this broader family of online flow-policy optimization methods. Finally, we validate ReFPO across GridWorld, MuJoCo Playground, and Humanoid Control benchmarks. ReFPO achieves high-quality one-step inference, with one-step scores matching or even exceeding multi-step results in several settings, while introducing no additional flow evaluations beyond the CFM computations already used in FPO.

\section{Related Work}

\subsection{Generative and Flow-based Policies in RL}
Early generative policies were primarily developed in the context of offline RL~\cite{janner2022planning,wangdiffusion,hansen2023idql,park2025flow,zhang2025energyweighted,wang2025one,seo2025fasttd3} to capture multimodal expert distributions. Recently, the field has increasingly shifted toward online reinforcement learning to achieve superior performance and environmental adaptability. Initial online methods, such as DDPO~\cite{black2024training}, DPPO~\cite{ren2025diffusion}, Flow-GRPO~\cite{liu2025flow}, and ReinFlow~\cite{zhangreinflow} typically treat the iterative denoising process as a multi-step Markov Decision Process (MDP), optimizing each discretized step via standard policy gradients. However, this MDP-based formulation departs from the continuous-time view of the underlying flows and suffers from high computational costs~\cite{li2025mixgrpo}.
 To address this, FPO~\cite{mcallister2025flow} introduced a holistic framework leveraging Conditional Flow Matching (CFM) as a variational lower bound~\cite{kingma2023understanding}. While FPO established a tractable gradient framework for flow-based policies, it primarily focuses on gradient directionality, leaving the evolution of intrinsic flow geometry across RL iterations underexplored--a gap that ReFPO aims to bridge.

\subsection{Path Rectification and Geometric Stability}
The sampling efficiency of continuous-time generative models is closely related to the straightness of their probability flows~\cite{shankar2025learning}. Early work on Rectified Flow and Reflow~\cite{liu2023flow} introduced recursive straightening procedures, and subsequent studies have developed objectives for learning more transport-efficient mappings~\cite{tong2024improving,kornilov2024optimal}. These advancements have progressively enabled one-step acceleration by unifying consistency-based~\cite{yang2024consistency,wu2025scot} objectives with path rectification. However, a persistent challenge in recursive straightening is the risk of model collapse~\cite{zhu2024analyzing}, where the vector field degenerates due to an over-reliance on self-consistency within self-generated data distributions. ReFPO addresses this by framing the policy optimization process as an implicit Reflow paradigm that incorporates a reward-driven corrective force. By synergizing the advantage-guided signal with the flow-matching objective, our method ensures that trajectory rectification is grounded in environmental feedback rather than mere self-imitation. This intrinsic synergy naturally reduces distributional collapse and maintains mode diversity without requiring additional training stages or auxiliary datasets.

\subsection{High-Fidelity One-Step Inference}
Achieving high-fidelity one-step inference has emerged as a pivotal research direction for overcoming the inference latency of generative models, spanning paradigms from supervised fine-tuning to reinforcement learning. In the supervised setting, various approaches improve one-step sampling by fitting straighter or more transport-efficient trajectories through specialized objectives~\cite{frans2025one,geng2025mean,luo2025soflow,zhang2025flow,liu2025flashaudio,zhou2025terminal}
, or by compressing pre-trained models via adversarial learning~\cite{cheng2025twinflow,lin2025adversarial} and knowledge distillation~\cite{zhu2025di,yin2024one}. Similarly, reinforcement learning methods aimed at one-step generation often rely on explicit teacher guidance~\cite{zhu2024slimflow,seo2025fasttd3}, computationally intensive reparameterized gradients~\cite{kornilov2024optimal,nguyen2025revisiting,lv2025flow,chen2025one,wang2025one}
, or additional training stages~\cite{park2025flow,li2025reinforcement}. Despite their progress in sampling speed, the common pitfalls of these methods lie in their substantial computational overhead or complex distillation pipelines, which significantly increase the threshold for practical implementation.

In contrast, ReFPO achieves high-fidelity one-step inference through a ``straightening-while-training'' paradigm. By integrating Reflow regularization into the RL objective, the probability flow is rectified during optimization, enabling $N=1$ sampling as a natural byproduct without auxiliary distillation or multi-stage pipelines.

\section{Reflow-regularized Flow Matching Policy Gradients}

\subsection{PPO and FPO}

% 内容： 简要回顾 PPO 的 Surrogate Objective 和 FPO 如何利用 CFM 变分下界替代似然率。这是为了统一定义符号（Notation）。
The objective of on-policy RL is to learn a policy \(\pi_\theta\) that maximizes expected return by increasing the likelihood of actions with positive advantage. Proximal Policy Optimization (PPO)~\cite{schulman2017proximal} stabilizes policy gradient updates by introducing a clipped likelihood ratio,
\begin{equation}
r(\theta)=\frac{\pi_\theta(a_t\mid o_t)}{\pi_{\theta_{\mathrm{old}}}(a_t\mid o_t)},
\end{equation}
and optimizing a clipped surrogate objective that constrains updates to remain within a local trust region. This formulation relies critically on access to tractable action likelihoods. However, exact action likelihoods are generally unavailable or expensive to compute for the flow-matching policies considered here.

FPO addresses this limitation by replacing PPO’s likelihood ratio with a computable surrogate derived from flow matching objectives. Instead of explicitly evaluating \(\pi_\theta(a_t\mid o_t)\), FPO constructs a likelihood-ratio proxy by comparing CFM losses under the current and old policies. Concretely, FPO defines a proxy ratio
\begin{equation}
\hat r^{\mathrm{FPO}}(\theta)
=
\exp\!\left(
\widehat{\mathcal L}^{\mathrm{CFM}}_{\theta_{\mathrm{old}}}(a_t;o_t)
-
\widehat{\mathcal L}^{\mathrm{CFM}}_{\theta}(a_t;o_t)
\right).
\end{equation}
FPO then substitutes \(\hat r^{\mathrm{FPO}}(\theta)\) directly into the PPO clipped surrogate objective.

This substitution is theoretically motivated by the connection between flow matching losses and the evidence lower bound (ELBO) of the induced policy distribution. Following FPO, this substitution is justified by the variational interpretation of flow matching: the weighted denoising/CFM loss estimates the negative ELBO up to a \(\theta\)-independent constant. Therefore, the exponential difference of CFM losses estimates the ratio of ELBOs under the old and current policies, serving as a tractable proxy for the likelihood ratio used in PPO.

With this replacement, FPO preserves the core structure of PPO, including clipping, advantage weighting, and on-policy updates, while enabling the training of expressive flow-based policies without explicit likelihood computation.

\subsection{FPO as an Implicit Online Reflow Process}

% 内容： 你的核心证明。阐述 FPO 在训练迭代中，其采样数据对 (a 0,a 1) 本质上是由旧策略生成的，这与 Rectified Flow 的数据生成逻辑一致。证明 FPO 在不断下降加权后的 Lcfm 过程中，潜移默化地在做路径拉直（Straightening）。
We show that, at the level of the unclipped policy-gradient surrogate, FPO admits an implicit advantage-weighted Reflow interpretation in an online, self-sampled setting, rather than merely inheriting the straightening property of CFM in supervised learning.

A key distinction from standard flow matching formulations is that, in FPO,
the action pairs \((\epsilon,a_1)\) used to define the CFM objective are not drawn from a fixed dataset. Instead, they are generated on-policy by the flow policy itself. As a result, the learning dynamics form a closed loop: the policy both induces the training distribution and is updated based on
it. This self-sampling property is precisely what connects FPO to Reflow, where rectification is applied recursively to model-generated samples.

To make this connection explicit, consider the unclipped FPO surrogate objective
\begin{equation}
\mathcal L(\theta)
=
-\mathbb{E}_t\big[\hat r_t(\theta) A_t\big],
\end{equation}
where the proxy ratio is defined as
\begin{equation}
\hat r_t(\theta)
=
\exp\!\Big(
\mathcal L^{\mathrm{CFM},\theta_{\mathrm{old}}}(a_t;o_t)
-
\mathcal L^{\mathrm{CFM},\theta}(a_t;o_t)
\Big).
\end{equation}
Since the old-policy loss \(\mathcal{L}^{\mathrm{CFM},\theta_{\mathrm{old}}}(a_t;o_t)\) is treated as a stop-gradient, taking the gradient with respect to \(\theta\) then yields
\begin{equation}
\begin{aligned}
\frac{\partial}{\partial\theta}\big(\hat r_t(\theta)A_t\big)
&= -A_t\hat r_t(\theta)\,
\nabla_\theta\mathcal{L}^{\mathrm{CFM},\theta}(a_t;o_t).
\end{aligned}
\end{equation}

For this unclipped surrogate, under gradient descent on the negative objective, positive advantages drive updates along $-\nabla_\theta \mathcal{L}^{\mathrm{CFM}}$, whereas negative advantages invert the gradient direction.

Furthermore, we unpack the CFM loss per action sample:
\begin{equation}
\mathcal{L}^{\mathrm{CFM},\theta}(a_t;o_t)
=\mathbb{E}_{\tau,\epsilon}\big[\|\,v_\theta(a_{\tau,t},\tau;o_t)-(a_t-\epsilon)\,\|^2\big],
\end{equation}

with noisy intermediate \(a_{\tau,t}=\alpha_\tau a_t+\sigma_\tau\epsilon\). Minimizing the per-sample CFM loss encourages the model velocity field \(v_\theta(\cdot,\tau)\) at every intermediate \(a_{\tau,t}\) to align with the same vector \(a_t-\epsilon\). Therefore, minimizing this loss encourages the velocity field along straight segments connecting self-generated noise–action pairs.

Taken together, these observations show that FPO performs an advantage-weighted, online Reflow process, in which the policy repeatedly rectifies its own induced trajectories under task-driven feedback. This perspective motivates ReFPO, which makes FPO's implicit, advantage-weighted rectified-flow behavior explicit through Reflow regularization.

\subsection{Enhancing Stability via Explicit Reflow Regularization}

ReFPO improves one-step inference and PPO-style training stability by adding a
simple geometric prior to the FPO objective. The key idea is to make the learned
velocity field closer to a straight rectified-flow path on the same frozen
on-policy samples used by FPO. This distinction is important: during an inner
optimization phase, the sampled actions and interpolants are fixed, so our
stability argument should be understood as a fixed-batch sensitivity analysis
rather than as a derivative through the data-collection process.

Given an on-policy sample \((o_t,a_t)\), we add the unweighted CFM regularizer
\begin{equation}
\mathcal L_{\mathrm{Reflow}}(\theta)
=
\mathbb E_{\tau,\epsilon}
\left[
\left\|
v_\theta(a_{\tau,t},\tau;o_t) - (a_t-\epsilon)
\right\|_2^2
\right],
\end{equation}
where \(a_{\tau,t}=\alpha_\tau a_t+\sigma_\tau\epsilon\). The full objective is
\begin{equation}
\mathcal L_{\mathrm{ReFPO}}(\theta)
=
\mathcal L_{\mathrm{FPO}}(\theta)
+
\lambda\,\mathcal L_{\mathrm{Reflow}}(\theta),
\end{equation}
The full training procedure is summarized in Algorithm~\ref{alg:refpo}
with \(\lambda>0\) controlling the strength of the auxiliary geometric term.
Unlike the FPO surrogate, \(\mathcal L_{\mathrm{Reflow}}\) is not reweighted by
the advantage. It therefore does not decide which trajectories should be
reinforced by reward; instead, it regularizes the geometry of the fitted flow on
the same trajectories.

\paragraph{Rollout-time effect.}
At inference time, the policy action is obtained by integrating the learned flow
ODE
\[
\dot x(\tau)=v_\theta(x(\tau),\tau;o).
\]
For a coarse one-step solver, the local truncation error depends on how quickly
the velocity changes along the trajectory. A Taylor expansion gives
\begin{equation}
\begin{aligned}
x(\tau-h)
&=
x(\tau)
-
h\,v_\theta(x(\tau),\tau;o)
+
\frac{h^2}{2}
\left(
\partial_\tau v_\theta
+
J_x v_\theta\,v_\theta
\right)_{(x(\tau),\tau;o)}+\mathcal O(h^3).
\end{aligned}
\end{equation}
Consequently, a large spatial Jacobian can amplify one-step discretization
error through the term \(J_x v_\theta\,v_\theta\):
\begin{equation}
\|\mathrm{LTE}\|
\lesssim
\frac{h^2}{2}
\left(
\|\partial_\tau v_\theta\|
+
\|J_x v_\theta\|_{\mathrm{op}}\|v_\theta\|
\right)
+
\mathcal O(h^3).
\end{equation}
By encouraging the fitted velocity to align with the straight direction
\(a_t-\epsilon\), the Reflow regularizer reduces unnecessary bending of the
learned field. This makes one-step rollout closer to multi-step integration,
without requiring a separate distillation or consistency-training stage.

\paragraph{Fixed-batch training effect.}
The same regularizer also makes the PPO-style inner loop less sensitive on a
frozen batch. Consider a fixed batch
\((o_t,a_t,\{\tau_i,\epsilon_i\}_{i=1}^N)\) and define
\begin{equation}
\begin{aligned}
\widehat L_{\mathrm{batch}}(\theta)
&=
\frac{1}{N}\sum_{i=1}^N
\left\|
v_\theta(z_i,\tau_i;o_t)-(a_t-\epsilon_i)
\right\|_2^2, \\
z_i
&:=
\alpha_{\tau_i}a_t+\sigma_{\tau_i}\epsilon_i .
\end{aligned}
\end{equation}
Here \(z_i\) is fixed when \(\theta\) is updated inside the inner loop. Thus,
the relevant question is how the fixed-batch surrogate changes as the current parameter moves from \(\theta_{\mathrm{old}}\) to \(\theta_{\mathrm{old}}+\delta\theta\)
.
A first-order expansion gives
\begin{equation}
\begin{aligned}
\widehat L_{\mathrm{batch}}(\theta_{\mathrm{old}}+\delta\theta)
&=
\widehat L_{\mathrm{batch}}(\theta_{\mathrm{old}})
+
\left\langle
\nabla_\theta \widehat L_{\mathrm{batch}}(\theta_{\mathrm{old}}),
\delta\theta
\right\rangle
+
\mathcal O(\|\delta\theta\|^2),
\end{aligned}
\end{equation}
where
\begin{equation}
\begin{aligned}
\nabla_\theta \widehat L_{\mathrm{batch}}(\theta_{\mathrm{old}})
&=
\frac{2}{N}\sum_{i=1}^N
\left(\partial_\theta
v_{\theta_{\mathrm{old}}}(z_i,\tau_i;o_t)\right)^\top \cdot
\left(
v_{\theta_{\mathrm{old}}}(z_i,\tau_i;o_t)
-
(a_t-\epsilon_i)
\right).
\end{aligned}
\end{equation}
The FPO proxy ratio can be written in terms of the fixed-batch CFM change as
\begin{equation}
\rho(\theta)
:=
\exp\!\left(
\widehat L_{\mathrm{batch}}(\theta_{\mathrm{old}})
-
\widehat L_{\mathrm{batch}}(\theta)
\right).
\end{equation}
Therefore,
\begin{equation}
\begin{aligned}
\log \rho(\theta_{\mathrm{old}}+\delta\theta)
&=
-
\left\langle
\nabla_\theta \widehat L_{\mathrm{batch}}(\theta_{\mathrm{old}}),
\delta\theta
\right\rangle 
+
\mathcal O(\|\delta\theta\|^2).
\end{aligned}
\end{equation}
This expression shows that ratio volatility is affected, to first order, by the
residual of the fitted velocity field and its parameter sensitivity on the
frozen batch. Reflow directly penalizes the same residual without advantage
reweighting and biases the model toward a simpler, straighter field. As a
result, the fixed-batch CFM proxy varies more smoothly across parameter updates,
which empirically reduces ratio spikes and the frequency of clipping events.

These two effects are complementary. The geometric regularizer improves
rollout-time one-step accuracy by reducing unnecessary curvature, and it
stabilizes the inner-loop optimization by reducing the first-order sensitivity
of the fixed-batch surrogate. Together, they provide a mechanism consistent with the observed improvements in both
one-step action quality and training stability in our experiments.

\begin{figure*}[t]
    \centering

    \newcommand{\imgheight}{2.75cm}

    % Column titles
    \begin{minipage}{0.05\textwidth}
        \hfill
    \end{minipage}%
    \hfill
    \begin{minipage}{0.22\textwidth}
        \centering
        \textbf{Vector (10-step)}
    \end{minipage}%
    \hfill
    \begin{minipage}{0.22\textwidth}
        \centering
        \textbf{Vector (1-step)}
    \end{minipage}%
    \hfill
    \begin{minipage}{0.22\textwidth}
        \centering
        \textbf{Traj. (10-step)}
    \end{minipage}%
    \hfill
    \begin{minipage}{0.22\textwidth}
        \centering
        \textbf{Traj. (1-step)}
    \end{minipage}

    \vspace{0.25em}

    % FPO row
    \begin{minipage}{0.05\textwidth}
        \centering
        \rotatebox{90}{\textbf{FPO}}
    \end{minipage}%
    \hfill
    \begin{minipage}{0.22\textwidth}
        \centering
        \includegraphics[width=\linewidth,height=\imgheight,keepaspectratio]{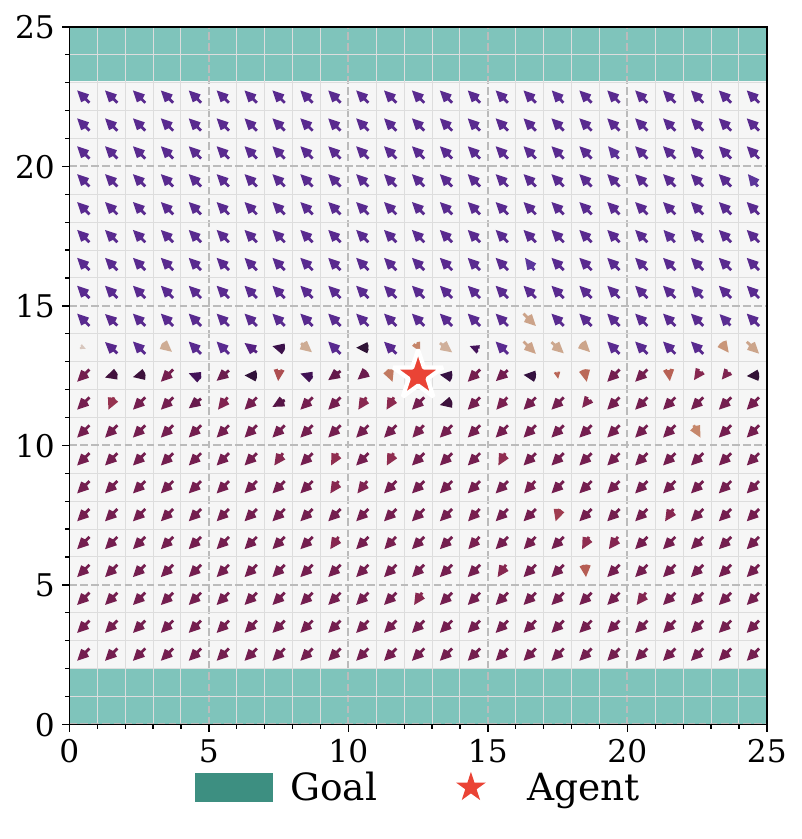}
    \end{minipage}%
    \hfill
    \begin{minipage}{0.22\textwidth}
        \centering
        \includegraphics[width=\linewidth,height=\imgheight,keepaspectratio]{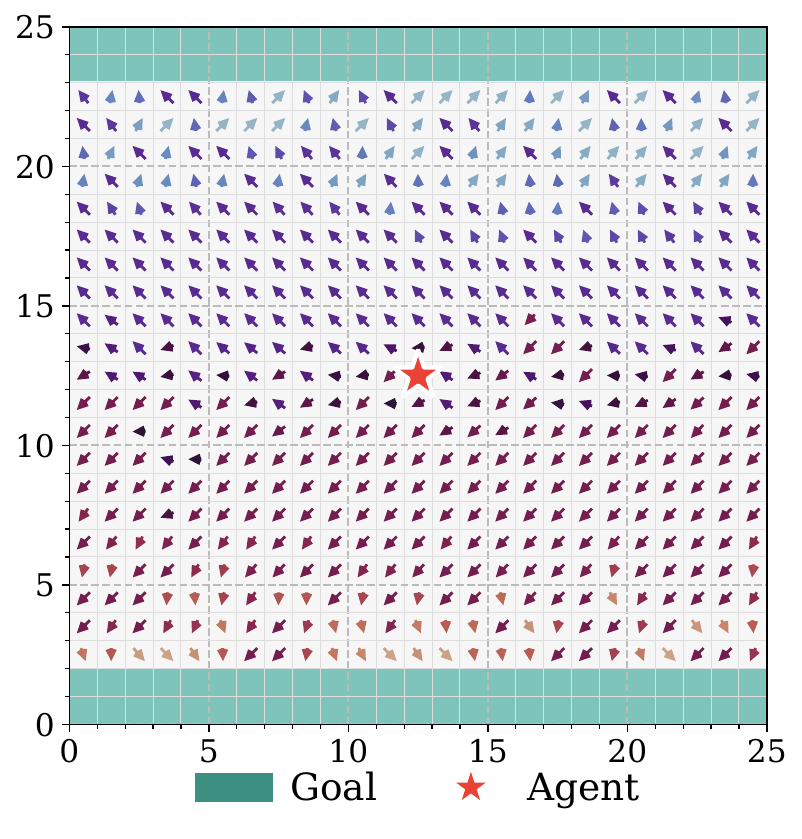}
    \end{minipage}%
    \hfill
    \begin{minipage}{0.22\textwidth}
        \centering
        \includegraphics[width=\linewidth,height=\imgheight,keepaspectratio]{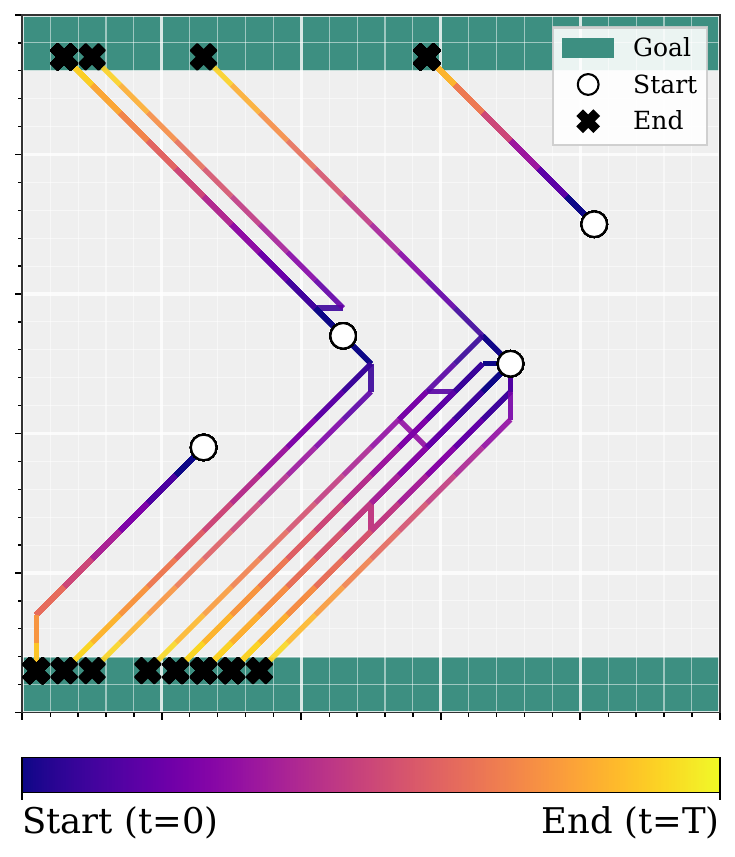}
    \end{minipage}%
    \hfill
    \begin{minipage}{0.22\textwidth}
        \centering
        \includegraphics[width=\linewidth,height=\imgheight,keepaspectratio]{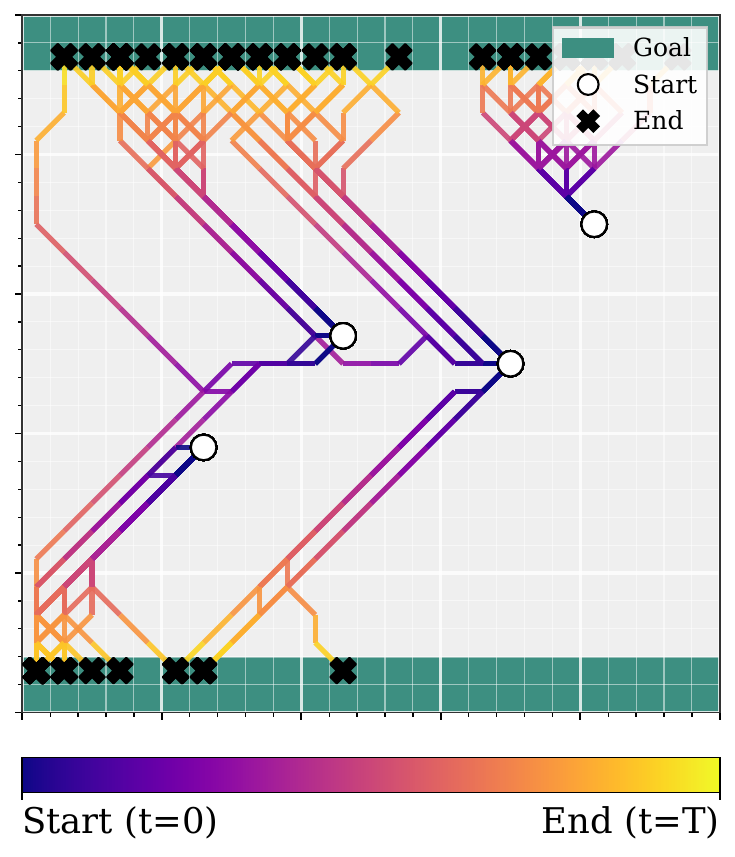}
    \end{minipage}

    \vspace{-0.2em}

    % ReFPO row
    \begin{minipage}{0.05\textwidth}
        \centering
        \rotatebox{90}{\textbf{ReFPO}}
    \end{minipage}%
    \hfill
    \begin{minipage}{0.22\textwidth}
        \centering
        \includegraphics[width=\linewidth,height=\imgheight,keepaspectratio]{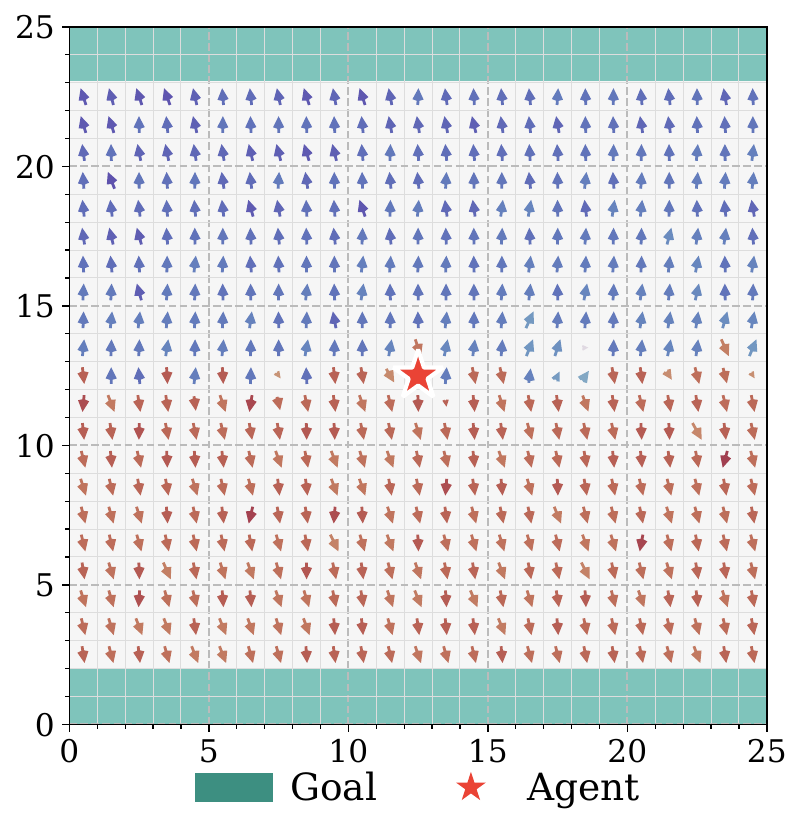}
    \end{minipage}%
    \hfill
    \begin{minipage}{0.22\textwidth}
        \centering
        \includegraphics[width=\linewidth,height=\imgheight,keepaspectratio]{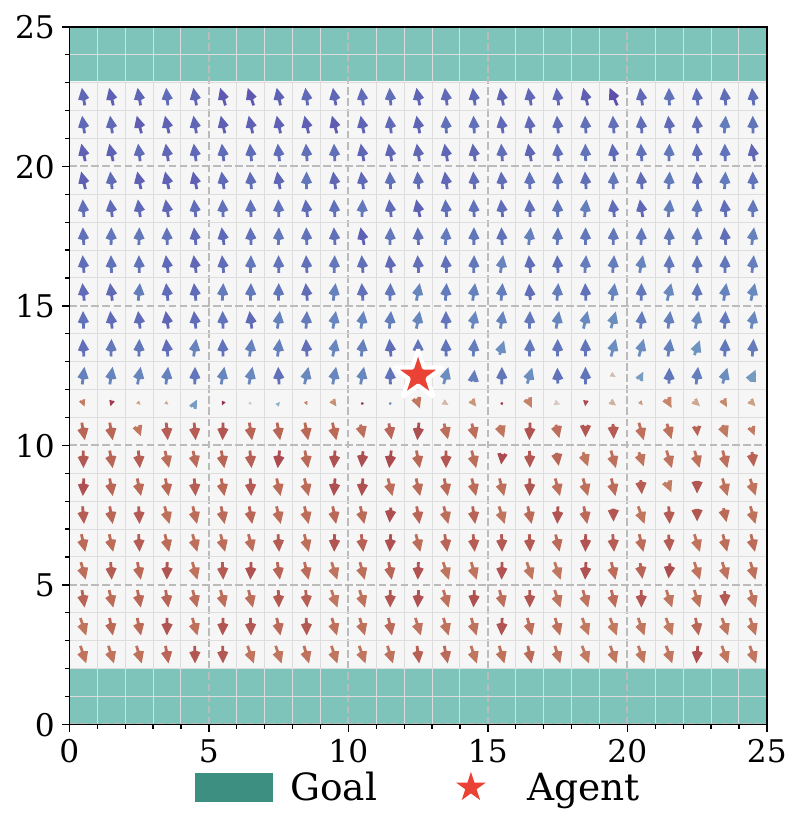}
    \end{minipage}%
    \hfill
    \begin{minipage}{0.22\textwidth}
        \centering
        \includegraphics[width=\linewidth,height=\imgheight,keepaspectratio]{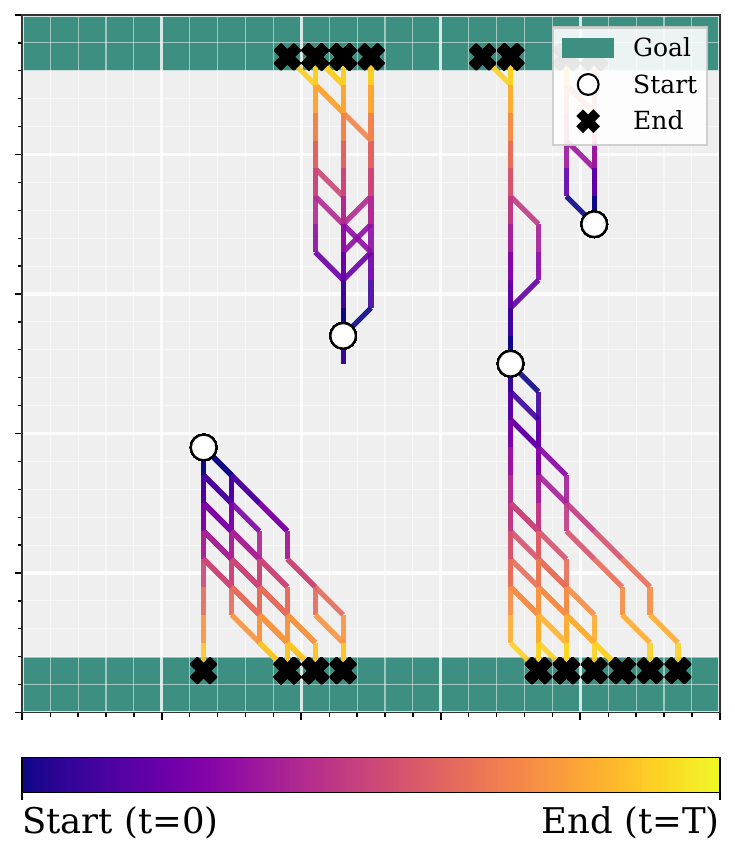}
    \end{minipage}%
    \hfill
    \begin{minipage}{0.22\textwidth}
        \centering
        \includegraphics[width=\linewidth,height=\imgheight,keepaspectratio]{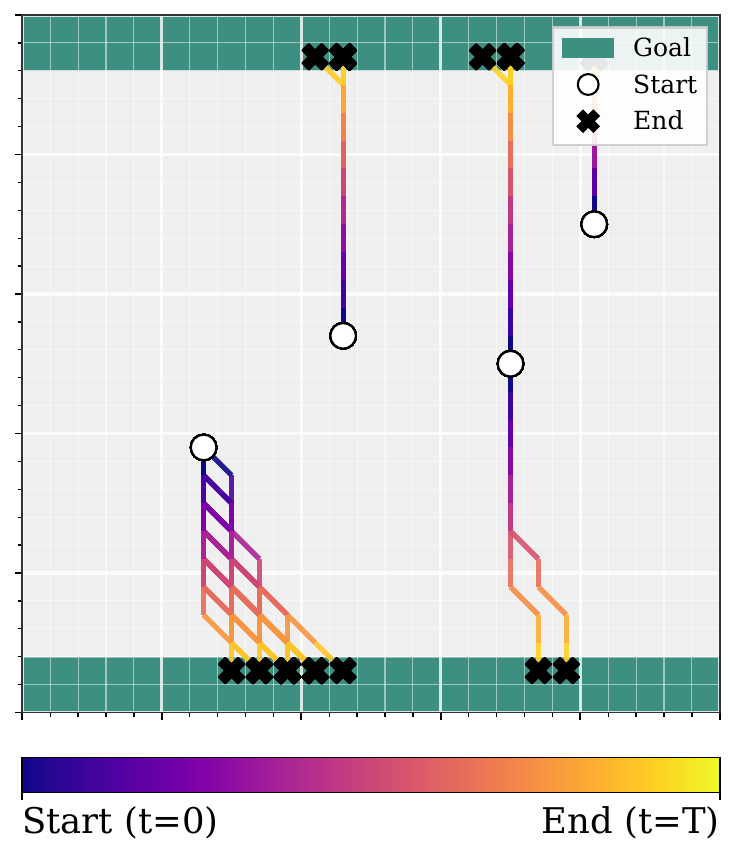}
    \end{minipage}

    \vspace{-0.3em}

    \caption{\textbf{Visualizations on the Multimodal Grid World task.} The top and bottom rows correspond to FPO and ReFPO, respectively. Left columns: learned vector fields representing the policy's action distribution. Right columns: rollout trajectories starting from fixed initial states to goal zones. Both visualizations are shown for 10-step and 1-step generation settings.}
    \label{fig:grid_comparison}
\end{figure*}

\section{Experiments}

We evaluate ReFPO across three distinct domains: a multi-goal Grid World to visualize distribution straightness~\cite{brockman2016openaigym,towers2024gymnasium}, MuJoCo Playground~\cite{todorov2012mujoco,zakka2025mujoco} for standard continuous control benchmarks, and high-dimensional Humanoid Control to test tracking under-conditioned control signals. Unless otherwise noted, quantitative results are averaged over five random seeds, and reported \(\pm\) values denote standard deviations across seeds.

% Our experiments aim to demonstrate three key properties of the proposed \textbf{ReFPO} algorithm:

% \begin{itemize}
%     \item \textbf{Inherent Reflow Property}: FPO’s iterative training on its own rollout data is equivalent to the "rectification" process.
%     \item \textbf{Straightness \& Efficiency}: ReFPO significantly reduces the discretization error, enabling high-performance policy execution with a single Euler step ($N=1$).
%     \item \textbf{Training Stability}: By augmenting the FPO objective with a Reflow regularization term, we mitigate trust-region instability and bootstrap errors.
% \end{itemize}

\subsection{GridWorld}

We evaluate ReFPO in a $25 \times 25$ GridWorld with bifurcated sparse rewards to test whether it preserves multimodal action distributions while improving discretization robustness. We compare against vanilla FPO~\cite{mcallister2025flow}.

\paragraph{Multimodality and Path Geometry.}
Figure~\ref{fig:grid_comparison} visualizes the denoising trajectories of FPO and ReFPO under $N=10$ Euler steps. ReFPO preserves the multimodal structure of the policy: trajectories still bifurcate near the central saddle point and reach both goal regions. Compared with FPO, ReFPO produces straighter and more spatially consistent trajectories, suggesting that the explicit Reflow regularizer reduces unnecessary curvature without collapsing the two modes.

\paragraph{One-Step Inference.}
Figure~\ref{fig:grid_comparison} also compares one-step ($N=1$) generation. Vanilla FPO can already complete this simple task with a single Euler step, suggesting that some path straightening emerges naturally during online flow-policy training. However, its one-step action field still exhibits local fluctuations and directional drift, which can make coarse discretization less reliable. ReFPO reduces these artifacts and produces more coherent one-step samples, indicating that explicit Reflow regularization improves discretization robustness without collapsing the multimodal structure of the policy.

% \begin{figure}[h]
%   \centering
%   \includegraphics[width=\textwidth]{figures/Playground.pdf}
%   \caption{
%   Comparison of FPO and ReFPO on DM Control Suite. The curves illustrate the evaluation rewards over 100M environment steps. The shaded regions represent the performance gap between the 10-step and 1-step generation for each method. A narrower shaded area indicates higher consistency between the accelerated 1-step policy and the standard 10-step baseline.
%   }
%   \label{figure:mujoco}
% \end{figure}
% \begin{table*}
%   \caption{\textbf{ReFPO variant comparison.} Cumulative rewards (mean $\pm$ std), straightness, and stability across different configurations. ReFPO* denotes the variant with $\lambda=0.04$.}
%   \label{table:mujoco}
%   \begin{center}
%     \begin{small}
%       \begin{sc}
%         \begin{tabular}{lcccc}
%           \toprule
%           Method  & Reward  & Reward(1step)& Straightness Error & Explosion Rate  \\
%           \midrule
%           PPO    & 622$\pm192$ &/ & / & /  \\
%           FPO & 641$\pm140$ & 565$\pm160$ & 0.0475 & 0.00365 \\
%           ReFPO*    & \textbf{686}$\pm\textbf{139}$& \textbf{690}$\pm\textbf{139}$& 0.0116 & \textbf{0.00189} \\
%             \midrule
%           ReFPO,$\lambda=0.02$& 644$\pm132$ & 556$\pm160$ & \textbf{0.00472} & 0.00377         \\
%           ReFPO,$\lambda=0.08$ &646$\pm141$ & 652$\pm151$ & 0.0161 & 0.00569 \\
%           ReFPO,$\lambda=0.12$& 609$\pm147$ & 601$\pm155$ & 0.0193 & 0.00330 \\
%           \bottomrule
%         \end{tabular}
%       \end{sc}
%     \end{small}
%   \end{center}
%   \vskip -0.1in
% \end{table*}

\subsection{MuJoCo Playground}
\begin{figure}[h]
  \centering
  \includegraphics[width=\textwidth]{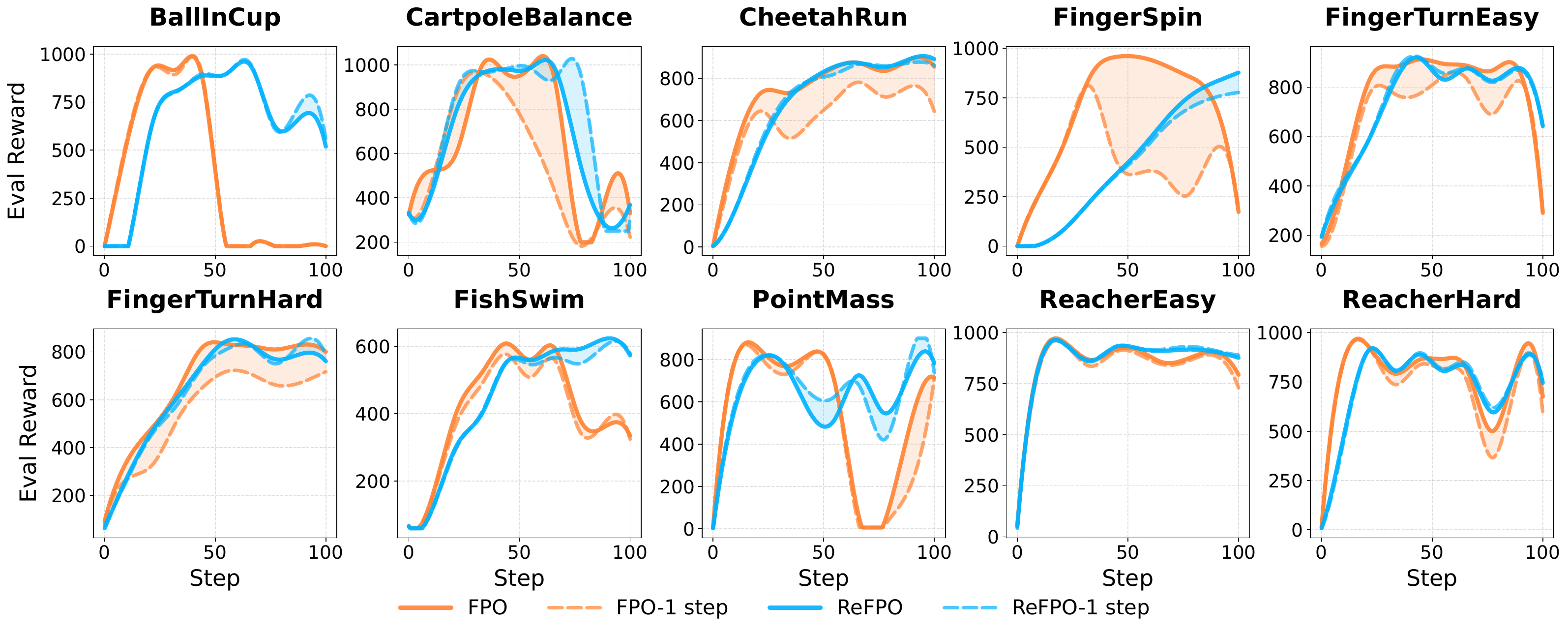}
\caption{
\textbf{MuJoCo Playground evaluation.}
FPO and ReFPO rewards on 10 DM Control Suite tasks over 100M steps. Shaded regions show the 10-step/1-step gap; narrower gaps indicate better one-step consistency.
}
  \label{figure:mujoco}
\end{figure}
\begin{figure}[h]
  \centering
  \includegraphics[width=\linewidth]{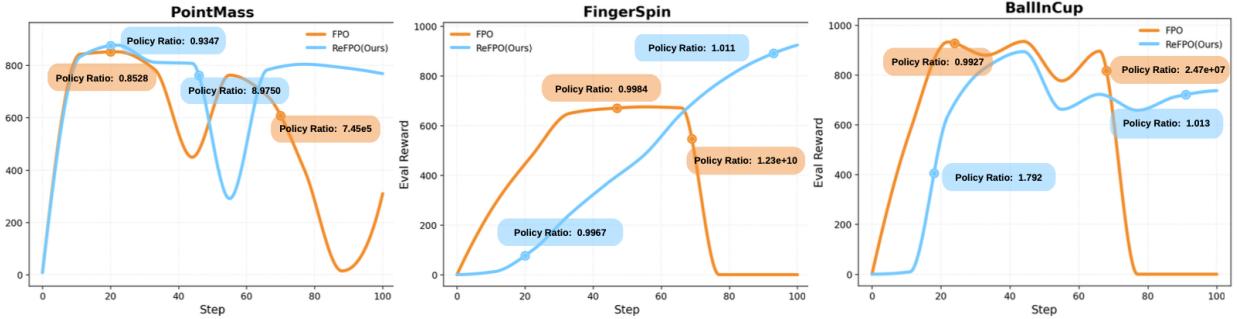}
\caption{
\textbf{Policy-ratio diagnostics.}
On PointMass, FingerSpin, and BallInCup, reward drops in FPO coincide with large proxy-ratio spikes, while ReFPO keeps the ratio smaller and the reward curves more stable.
}
  \label{fig:policy_ratio_diagnostics}
\end{figure}

We evaluate ReFPO on 10 continuous-control tasks adapted from the DeepMind Control Suite~\cite{tassa2018deepmind,tunyasuvunakool2020dm_control} within MuJoCo Playground, testing standard online control performance and low-latency one-step inference.

\paragraph{Experimental Setup and Metrics.}
Following FPO~\cite{mcallister2025flow}, we use Adam, batch size 1024, and 16 updates per batch. All train-from-scratch methods are trained for 100M environment steps and evaluated with $N=1$ and $N=10$ inference steps. Since online RL for flow-matching policies is recent and directly comparable baselines remain limited, we use FPO as the closest likelihood-free flow-policy baseline and include Gaussian PPO as a strong non-generative control baseline. We further compare with ReinFlow and Flow-GRPO, two state-of-the-art flow-matching post-training baselines, using the same tasks and tuning budget; we report their best tuned configurations after an additional 100M continued-training steps, with details in Section~\ref{appendix:mujoco}.

% \begin{table}[h]
% \centering
% \caption{
% \textbf{MuJoCo Playground results.}
% Mean reward and flow diagnostics across 10 tasks. ReFPO* denotes $\lambda=0.04$.
% }
% \label{tab:mujoco}
% \small
% \setlength{\tabcolsep}{5pt}

% % \begin{tabular}{lcccc}
% \begin{tabular}{p{2.8cm}cccc}
% \toprule
% Method & 10-step Reward & 1-step Reward & Straightness & Explosion \\
% \midrule
% \multicolumn{5}{l}{\textit{Training stage: 0--100m steps}} \\
% PPO & $622 \pm 192$ & -- & -- & -- \\
% FPO & $641 \pm 140$ & $565 \pm 160$ & 0.0475 & 0.00365 \\

% ReinFlow & $416 \pm 195$ & $339 \pm 194$ & -- & -- \\
% GRPO & $475 \pm 194$ & $0.000144 \pm 0.000218$ & -- & -- \\
% Trajectory-advantage-weighted ReFPO & $671.12 \pm 160.8$ & $639.52 \pm 165.74$ & -- & -- \\

% ReFPO* & $\mathbf{686 \pm 139}$ & $\mathbf{690 \pm 139}$ & 0.0116 & \textbf{0.00189} \\
% ReFPO, $\lambda=0.02$ & $644 \pm 132$ & $556 \pm 160$ & \textbf{0.00472} & 0.00377 \\
% ReFPO, $\lambda=0.08$ & $646 \pm 141$ & $652 \pm 151$ & 0.0161 & 0.00569 \\
% ReFPO, $\lambda=0.12$ & $609 \pm 147$ & $601 \pm 155$ & 0.0193 & 0.00330 \\

% \midrule
% \multicolumn{5}{l}{\textit{Training stage: 100m--200m steps}} \\
% ReinFlow & $523 \pm 188$ & $357 \pm 178$ & -- & -- \\
% GRPO & $507 \pm 235$ & $0.000150 \pm 0.000226$ & -- & -- \\
% \bottomrule
% \end{tabular}
% \end{table}
\begin{table}[h]
\caption{
\textbf{MuJoCo Playground results.}
Mean reward and flow diagnostics across 10 tasks. ReFPO* denotes $\lambda=0.04$.
}
\label{tab:mujoco}
\centering
\small
\setlength{\tabcolsep}{4pt}
\renewcommand{\arraystretch}{1.12}

\begin{tabularx}{\linewidth}{Xcccc}
\toprule
\textbf{Method} & \textbf{10-step Reward} & \textbf{1-step Reward} & \textbf{Straightness} & \textbf{Explosion} \\
\midrule
\multicolumn{5}{@{}l}{\textit{Training stage: 0--100M steps}} \\
\addlinespace[2pt]
PPO & $622 \pm 192$ & -- & -- & -- \\
Flow-GRPO & $475 \pm 194$ & $0.000144 \pm 0.000218$ & -- & -- \\
ReinFlow & $416 \pm 195$ & $339 \pm 194$ & -- & -- \\
FPO & $641 \pm 140$ & $565 \pm 160$ & 0.0475 & 0.00365 \\
ReFPO* & $\mathbf{686 \pm 139}$ & $\mathbf{690 \pm 139}$ & \textbf{0.0116} & \textbf{0.00189} \\
\midrule
ReFPO, $\lambda=0.02$ & $644 \pm 132$ & $556 \pm 160$ & 0.0187 & 0.00377 \\
ReFPO, $\lambda=0.08$ & $646 \pm 141$ & $652 \pm 151$ & 0.0161 & 0.00569 \\
ReFPO, $\lambda=0.12$ & $609 \pm 147$ & $601 \pm 155$ & 0.0193 & 0.00330 \\
\makecell[l]{ReFPO,~Trajectory-advantage-\\weighted} & $671 \pm 161$ & $640 \pm 166$ & -- & -- \\

\midrule
\multicolumn{5}{@{}l}{\textit{Training stage: 100M--200M steps}} \\
\addlinespace[2pt]
Flow-GRPO  & $507 \pm 235$ & $0.000150 \pm 0.000226$ & -- & -- \\
ReinFlow & $523 \pm 188$ & $357 \pm 178$ & -- & -- \\
\bottomrule
\end{tabularx}
\end{table}

Beyond cumulative reward, we report two flow diagnostics. Straightness Error is the MSE between the learned velocity field $v_\theta$ and the straight target direction $(a_1-a_0)$; lower values indicate a more linear sampling path and smaller expected discretization error. Explosion Rate is the fraction of samples with $|\mathcal{L}_{\text{new}}-\mathcal{L}_{\text{old}}|>3$, corresponding to a proxy-ratio change larger than $\exp(3)$; it captures large numerical spikes in the exponential proxy ratio. Across tested coefficients, $\lambda=0.04$ gives the best reward--stability trade-off.

\paragraph{Results and Dynamics Analysis.}
Figure~\ref{fig:policy_ratio_diagnostics} provides a targeted diagnostic on PointMass, FingerSpin, and BallInCup, where vanilla FPO exhibits visible reward degradation. In these tasks, large CFM proxy-ratio spikes consistently appear before the reward drops of FPO. Since the log proxy ratio is the difference between the old and new CFM losses, these spikes indicate abrupt changes in the same CFM quantity used by the PPO-style clipping surrogate. ReFPO does not directly minimize the proxy ratio. Instead, its straightening regularizer acts on the same flow-matching geometry, making the CFM proxy more stable in practice and substantially reducing the frequency of extreme ratio events.

The full 10-task curves in Figure~\ref{figure:mujoco} show the benchmark-level effect. ReFPO achieves higher average reward than FPO and substantially reduces the gap between 10-step and 1-step evaluation, as reflected by the smaller shaded regions. Interestingly, in several tasks one-step evaluation is comparable to or slightly better than 10-step evaluation, which we attribute to imperfect learned fields where multi-step rollout can accumulate intermediate velocity errors.

Table~\ref{tab:mujoco} provides the corresponding summary: on average, ReFPO improves both 10-step and 1-step rewards while lowering Straightness Error and Explosion Rate. The post-training baselines remain below ReFPO even with additional continued training: ReinFlow retains one-step capability but obtains lower returns, while Flow-GRPO shows severe one-step degradation under coarse discretization. ReFPO also outperforms Gaussian PPO on average while supporting low-latency one-step flow inference.

Finally, the coefficient ablation shows that $\lambda=0.04$ gives the best reward--stability trade-off. To test whether the improvement comes merely from further emphasizing high-advantage samples, we also evaluate an advantage-weighted straightening variant, where the Reflow loss is applied preferentially to positive-advantage samples. This variant remains below standard ReFPO, especially in one-step evaluation, suggesting that consistent geometry regularization across sampled trajectories is more effective than selectively straightening only reward-favored directions.

\subsection{Humanoid Control}

\begin{figure}[h]
  \centering

  \begin{subfigure}[b]{0.65\textwidth}
    \centering
    \includegraphics[width=\textwidth]{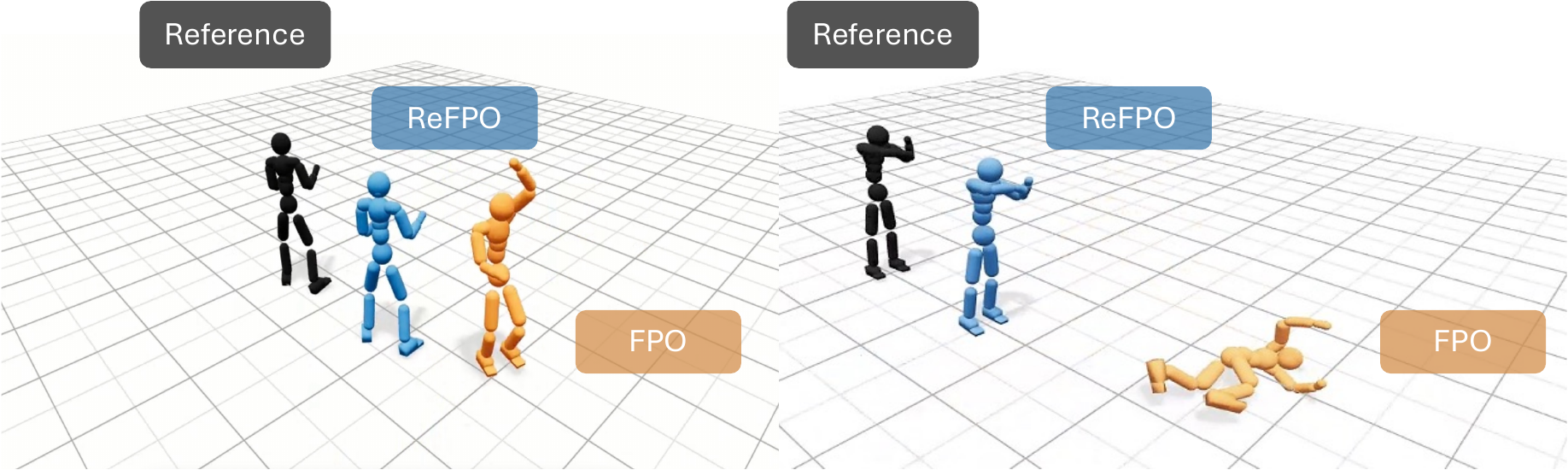}
    \caption{Performance on Humanoid Control.}
    \label{figure:humanoid}
  \end{subfigure}
  \hfill
  \begin{subfigure}[b]{0.25\textwidth}
    \centering
    \includegraphics[width=\textwidth]{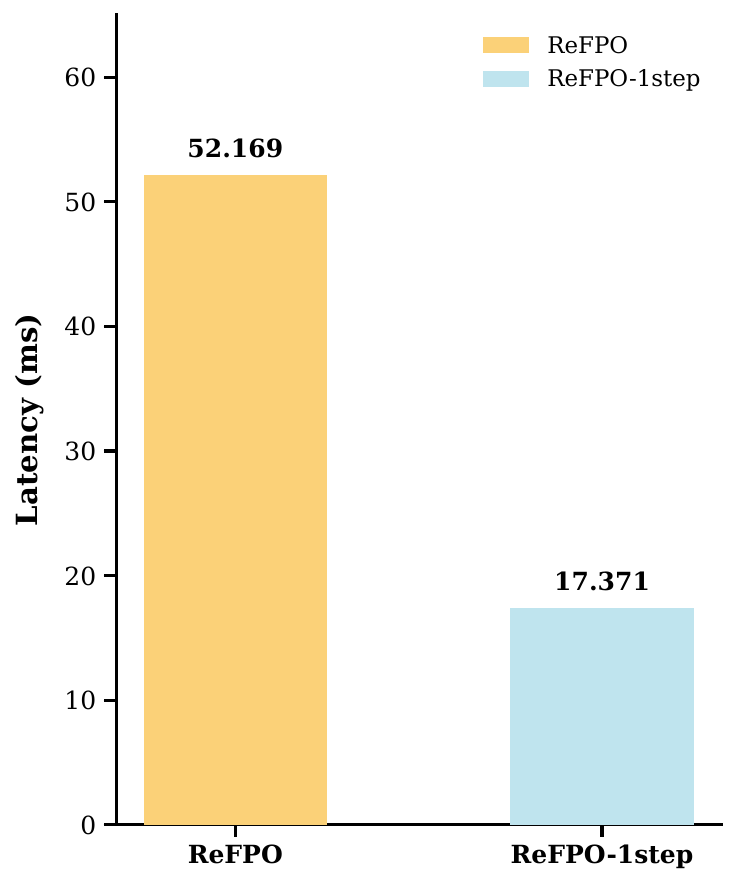}
    \caption{Inference latency comparison.}
    \label{fig:latency_bar}
  \end{subfigure}

  \caption{
    \textbf{Performance and efficiency analysis.}
    (a) Comparison of FPO and ReFPO on Humanoid Control.
    (b) Action generation latency, where ReFPO-1step achieves significantly lower inference time.
  }
  \label{fig:combined_results}
\end{figure}

Physics-aware humanoid control provides a high-dimensional benchmark for evaluating the scalability of flow-based policies. We consider a MoCap tracking task in which a simulated SMPL-based agent must reproduce diverse AMASS reference motions~\cite{mahmood2019amass} while maintaining physical balance and coordinated full-body control. This setting stresses both the expressivity of the action distribution and the discretization robustness required for low-latency inference. As shown in Figure~\ref{figure:humanoid}, ReFPO tracks the reference motion more stably than FPO, with additional examples provided in the supplementary materials.

\paragraph{Implementation Details and Metrics.}
The agent has 24 actuated joints (72 DoF) and is trained in Isaac Gym~\cite{makoviychuk2021isaac} using Puffer-PHC~\cite{luo2023perpetual}. Policies receive proprioception and goal-conditioned targets under three conditioning regimes: full-joint, root+hands, and root-only; in the latter two sparse-conditioning settings, multimodal action modeling becomes more important. All policies are trained for 100M environment steps with the imitation reward from~\cite{peng2018deepmimic}.

\begin{table*}[h]
\caption{\textbf{Quantitative evaluation on Humanoid Control tasks.} We report performance across three levels of conditioning difficulty using Success Rate ($\uparrow$), Alive Duration ($\uparrow$), and Mean Per Joint Position Error (MPJPE, $\downarrow$).}
\label{table:humanoid_results}
\begin{center}
\begin{small}
\begin{sc}

\begin{tabular}{ll|ccc}
\toprule
Methods & Goal conditioning & Success rate ($\uparrow$) & Alive duration ($\uparrow$) & MPJPE ($\downarrow$)  \\
\midrule
Gaussian PPO & All joints   & \textbf{98.7}\% & \textbf{200.46} & \textbf{31.62}       \\
FPO          & All joints   & 96.2\% & 197.64 &  42.65 \\
FPO-1step    & All joints   & 96.9\% & 198.25 &  41.20 \\
ReFPO*       & All joints   & 96.6\% & 198.03 &  41.00\\
ReFPO*-1step & All joints   & 97.3\% & 198.47 &  39.49\\
\midrule
Gaussian PPO & Root + Hands & 46.5\% & 142.50 & 97.65         \\
FPO          & Root + Hands & 71.0\% & 171.68&  63.88 \\
FPO-1step    & Root + Hands & 70.8\% & 171.37 &  64.96 \\
ReFPO*       & Root + Hands & 72.7\% & \textbf{174.98} &  62.68 \\
ReFPO*-1step & Root + Hands & \textbf{74.4}\% & 174.83 &  \textbf{60.60} \\
\midrule
Gaussian PPO & Root         & 29.8\% & 114.06 & 123.70         \\
FPO          & Root         & 55.0\% & 147.94 & 73.55\\
FPO-1step    & Root         & 55.3\% & 147.90 & 72.65 \\
ReFPO*       & Root         & 58.2\% &   156.37  & 70.97 \\
ReFPO*-1step & Root         & \textbf{58.7}\% &    \textbf{156.45}  & \textbf{70.89} \\
\bottomrule
\end{tabular}
\end{sc}
\end{small}
\end{center}
\vskip -0.1in
\end{table*}

\paragraph{Results and Qualitative Analysis.}
Table~\ref{table:humanoid_results} shows that ReFPO maintains strong one-step performance in high-dimensional humanoid control. Across the flow-policy variants, the $N=1$ results are consistently comparable to, and sometimes better than, their $N=10$ counterparts, while Figure~\ref{fig:combined_results} shows the expected latency reduction from one-step action generation. This supports the practical motivation of ReFPO: high-quality actions can be generated with substantially lower inference cost.

The results also clarify where expressive flow policies are most useful. Under full-joint conditioning, Gaussian PPO remains the strongest method, indicating that a simple Gaussian policy is highly competitive when the target information is dense. As conditioning becomes sparser, however, the advantage shifts toward flow-based policies: FPO and ReFPO substantially outperform Gaussian PPO in the root+hands and root-only settings. ReFPO further improves over FPO in these sparse regimes, with its one-step variant achieving the best success rate and MPJPE. These results suggest that ReFPO is especially beneficial when incomplete goal information makes multimodal action modeling and robust one-step inference important.

\section{Discussion and Limitations}

ReFPO provides a simple geometric regularizer for online reward-weighted flow-matching optimization. While our implementation builds on FPO, the idea is broader than a single PPO-style surrogate: CFM-based online RL or fine-tuning methods, such as ORW-CFM-W2~\cite{fan2025online}, also optimize flow models through reward-weighted objectives on generated samples. Because the regularizer only reuses the CFM residual already computed by these methods, it may transfer to this broader family with little algorithmic overhead.

The current study also leaves several directions open. Our analysis focuses on local fixed-batch sensitivity within PPO-style inner updates, rather than a complete characterization of online RL dynamics. Our experiments are centered on FPO-style flow policies; extending the same principle to broader diffusion- or flow-policy RL methods is an important next step. The benefits are most pronounced in settings involving sparse conditioning, multimodal action distributions, or low-latency one-step inference, while Gaussian PPO remains strong in fully observed humanoid tracking. Finally, the Reflow coefficient is selected empirically, and adaptive coefficient schedules may further improve robustness.

\clearpage
\bibliographystyle{plainnat}
\bibliography{references}

@article{chi2025diffusion,
  author =        {Chi, Cheng and Xu, Zhenjia and Feng, Siyuan and
                   Cousineau, Eric and Du, Yilun and Burchfiel, Benjamin and
                   Tedrake, Russ and Song, Shuran},
  journal =       {The International Journal of Robotics Research},
  number =        {10-11},
  pages =         {1684--1704},
  publisher =     {Sage Publications Sage UK: London, England},
  title =         {Diffusion policy: Visuomotor policy learning via
                   action diffusion},
  volume =        {44},
  year =          {2025},
}

@inproceedings{liu2023flow,
  author =        {Xingchao Liu and Chengyue Gong and qiang liu},
  booktitle =     {The Eleventh International Conference on Learning
                   Representations},
  title =         {Flow Straight and Fast: Learning to Generate and
                   Transfer Data with Rectified Flow},
  year =          {2023},
  url =           {https://openreview.net/forum?id=XVjTT1nw5z},
}

@misc{song2023consistencymodels,
  author =        {Yang Song and Prafulla Dhariwal and Mark Chen and
                   Ilya Sutskever},
  title =         {Consistency Models},
  year =          {2023},
  url =           {https://arxiv.org/abs/2303.01469},
}

@inproceedings{yin2024one,
  author =        {Yin, Tianwei and Gharbi, Micha{\"e}l and
                   Zhang, Richard and Shechtman, Eli and Durand, Fredo and
                   Freeman, William T and Park, Taesung},
  booktitle =     {Proceedings of the IEEE/CVF conference on computer
                   vision and pattern recognition},
  pages =         {6613--6623},
  title =         {One-step diffusion with distribution matching
                   distillation},
  year =          {2024},
}

@inproceedings{janner2022planning,
  author =        {Janner, Michael and Du, Yilun and Tenenbaum, Joshua and
                   Levine, Sergey},
  booktitle =     {International Conference on Machine Learning},
  organization =  {PMLR},
  pages =         {9902--9915},
  title =         {Planning with Diffusion for Flexible Behavior
                   Synthesis},
  year =          {2022},
}

@inproceedings{wangdiffusion,
  author =        {Wang, Zhendong and Hunt, Jonathan J and
                   Zhou, Mingyuan},
  booktitle =     {The Eleventh International Conference on Learning
                   Representations},
  title =         {Diffusion Policies as an Expressive Policy Class for
                   Offline Reinforcement Learning},
  year =          {2023},
}

@article{hansen2023idql,
  author =        {Hansen-Estruch, Philippe and Kostrikov, Ilya and
                   Janner, Michael and Kuba, Jakub Grudzien and
                   Levine, Sergey},
  journal =       {arXiv preprint arXiv:2304.10573},
  title =         {Idql: Implicit q-learning as an actor-critic method
                   with diffusion policies},
  year =          {2023},
}

@article{park2025flow,
  author =        {Park, Seohong and Li, Qiyang and Levine, Sergey},
  journal =       {arXiv preprint arXiv:2502.02538},
  title =         {Flow q-learning},
  year =          {2025},
}

@inproceedings{zhang2025energyweighted,
  author =        {Shiyuan Zhang and Weitong Zhang and Quanquan Gu},
  booktitle =     {The Thirteenth International Conference on Learning
                   Representations},
  title =         {Energy-Weighted Flow Matching for Offline
                   Reinforcement Learning},
  year =          {2025},
  url =           {https://openreview.net/forum?id=HA0oLUvuGI},
}

@article{wang2025one,
  author =        {Wang, Zeyuan and Li, Da and Chen, Yulin and Shi, Ye and
                   Bai, Liang and Yu, Tianyuan and Fu, Yanwei},
  journal =       {arXiv preprint arXiv:2511.13035},
  title =         {One-Step Generative Policies with Q-Learning: A
                   Reformulation of MeanFlow},
  year =          {2025},
}

@article{seo2025fasttd3,
  author =        {Seo, Younggyo and Sferrazza, Carmelo and Geng, Haoran and
                   Nauman, Michal and Yin, Zhao-Heng and Abbeel, Pieter},
  journal =       {arXiv preprint arXiv:2505.22642},
  title =         {FastTD3: Simple, Fast, and Capable Reinforcement
                   Learning for Humanoid Control},
  year =          {2025},
}

@inproceedings{black2024training,
  author =        {Kevin Black and Michael Janner and Yilun Du and
                   Ilya Kostrikov and Sergey Levine},
  booktitle =     {The Twelfth International Conference on Learning
                   Representations},
  title =         {Training Diffusion Models with Reinforcement
                   Learning},
  year =          {2024},
  url =           {https://openreview.net/forum?id=YCWjhGrJFD},
}

@inproceedings{ren2025diffusion,
  author =        {Allen Z. Ren and Justin Lidard and Lars Lien Ankile and
                   Anthony Simeonov and Pulkit Agrawal and
                   Anirudha Majumdar and Benjamin Burchfiel and
                   Hongkai Dai and Max Simchowitz},
  booktitle =     {The Thirteenth International Conference on Learning
                   Representations},
  title =         {Diffusion Policy Policy Optimization},
  year =          {2025},
  url =           {https://openreview.net/forum?id=mEpqHvbD2h},
}

@article{liu2025flow,
  author =        {Liu, Jie and Liu, Gongye and Liang, Jiajun and
                   Li, Yangguang and Liu, Jiaheng and Wang, Xintao and
                   Wan, Pengfei and Zhang, Di and Ouyang, Wanli},
  journal =       {arXiv preprint arXiv:2505.05470},
  title =         {Flow-grpo: Training flow matching models via online
                   rl},
  year =          {2025},
}

@inproceedings{zhangreinflow,
  author =        {Zhang, Tonghe and Yu, Chao and Su, Sichang and
                   Wang, Yu},
  booktitle =     {The Thirty-ninth Annual Conference on Neural
                   Information Processing Systems},
  title =         {ReinFlow: Fine-tuning Flow Matching Policy with
                   Online Reinforcement Learning},
  year =          {2025},
}

@article{li2025mixgrpo,
  author =        {Li, Junzhe and Cui, Yutao and Huang, Tao and
                   Ma, Yinping and Fan, Chun and Yang, Miles and
                   Zhong, Zhao},
  journal =       {arXiv preprint arXiv:2507.21802},
  title =         {Mixgrpo: Unlocking flow-based grpo efficiency with
                   mixed ode-sde},
  year =          {2025},
}

@article{mcallister2025flow,
  author =        {McAllister, David and Ge, Songwei and Yi, Brent and
                   Kim, Chung Min and Weber, Ethan and Choi, Hongsuk and
                   Feng, Haiwen and Kanazawa, Angjoo},
  journal =       {arXiv preprint arXiv:2507.21053},
  title =         {Flow matching policy gradients},
  year =          {2025},
}

@inproceedings{kingma2023understanding,
  author =        {Diederik P Kingma and Ruiqi Gao},
  booktitle =     {Thirty-seventh Conference on Neural Information
                   Processing Systems},
  title =         {Understanding Diffusion Objectives as the {ELBO} with
                   Simple Data Augmentation},
  year =          {2023},
  url =           {https://openreview.net/forum?id=NnMEadcdyD},
}

@inproceedings{shankar2025learning,
  author =        {Shiv Shankar and Tomas Geffner},
  booktitle =     {ICLR 2025 Workshop on Deep Generative Model in
                   Machine Learning: Theory, Principle and Efficacy},
  title =         {{LEARNING} {STRAIGHT} {FLOWS} {BY} {LEARNING}
                   {CURVED} {INTERPOLANTS}},
  year =          {2025},
  url =           {https://openreview.net/forum?id=9bJ2PJFNX4},
}

@article{tong2024improving,
  author =        {Tong, Alexander and Fatras, Kilian and
                   Malkin, Nikolay and Huguet, Guillaume and
                   Zhang, Yanlei and Rector-Brooks, Jarrid and Wolf, Guy and
                   Bengio, Yoshua},
  journal =       {Transactions on Machine Learning Research},
  pages =         {1--34},
  title =         {Improving and generalizing flow-based generative
                   models with minibatch optimal transport},
  year =          {2024},
}

@article{kornilov2024optimal,
  author =        {Kornilov, Nikita and Mokrov, Petr and
                   Gasnikov, Alexander and Korotin, Aleksandr},
  journal =       {Advances in Neural Information Processing Systems},
  pages =         {104180--104204},
  title =         {Optimal flow matching: Learning straight trajectories
                   in just one step},
  volume =        {37},
  year =          {2024},
}

@article{yang2024consistency,
  author =        {Yang, Ling and Zhang, Zixiang and Zhang, Zhilong and
                   Liu, Xingchao and Xu, Minkai and Zhang, Wentao and
                   Meng, Chenlin and Ermon, Stefano and Cui, Bin},
  journal =       {CoRR},
  title =         {Consistency Flow Matching: Defining Straight Flows
                   with Velocity Consistency},
  year =          {2024},
}

@inproceedings{wu2025scot,
  author =        {zhangkai wu and Xuhui Fan and Hongyu Wu and
                   Longbing Cao},
  booktitle =     {The Thirty-ninth Annual Conference on Neural
                   Information Processing Systems},
  title =         {{SC}oT: Unifying Consistency Models and Rectified
                   Flows via Straight-Consistent Trajectories},
  year =          {2025},
  url =           {https://openreview.net/forum?id=GV82iAD70j},
}

@article{zhu2024analyzing,
  author =        {Zhu, Huminhao and Wang, Fangyikang and Ding, Tianyu and
                   Qu, Qing and Zhu, Zhihui},
  journal =       {arXiv preprint arXiv:2412.08175},
  title =         {Analyzing and Mitigating Model Collapse in Rectified
                   Flow Models},
  year =          {2024},
}

@inproceedings{frans2025one,
  author =        {Kevin Frans and Danijar Hafner and Sergey Levine and
                   Pieter Abbeel},
  booktitle =     {The Thirteenth International Conference on Learning
                   Representations},
  title =         {One Step Diffusion via Shortcut Models},
  year =          {2025},
  url =           {https://openreview.net/forum?id=OlzB6LnXcS},
}

@article{geng2025mean,
  author =        {Geng, Zhengyang and Deng, Mingyang and Bai, Xingjian and
                   Kolter, J Zico and He, Kaiming},
  journal =       {arXiv preprint arXiv:2505.13447},
  title =         {Mean flows for one-step generative modeling},
  year =          {2025},
}

@article{luo2025soflow,
  author =        {Luo, Tianze and Yuan, Haotian and Liu, Zhuang},
  journal =       {arXiv preprint arXiv:2512.15657},
  title =         {SoFlow: Solution Flow Models for One-Step Generative
                   Modeling},
  year =          {2025},
}

@article{zhang2025flow,
  author =        {Zhang, Xinxi and Tan, Shiwei and Nguyen, Quang and
                   Dao, Quan and Han, Ligong and He, Xiaoxiao and
                   Zhang, Tunyu and Mrdovic, Alen and Metaxas, Dimitris},
  journal =       {arXiv preprint arXiv:2511.23342},
  title =         {Flow Straighter and Faster: Efficient One-Step
                   Generative Modeling via MeanFlow on Rectified
                   Trajectories},
  year =          {2025},
}

@inproceedings{liu2025flashaudio,
  author =        {Liu, Huadai and Wang, Jialei and Huang, Rongjie and
                   Liu, Yang and Lu, Heng and Zhao, Zhou and Xue, Wei},
  booktitle =     {Proceedings of the 63rd Annual Meeting of the
                   Association for Computational Linguistics (Volume 1:
                   Long Papers)},
  pages =         {13694--13710},
  title =         {FlashAudio: Rectified Flow for Fast and High-Fidelity
                   Text-to-Audio Generation},
  year =          {2025},
}

@article{zhou2025terminal,
  author =        {Zhou, Linqi and Parger, Mathias and Haque, Ayaan and
                   Song, Jiaming},
  journal =       {arXiv preprint arXiv:2511.19797},
  title =         {Terminal Velocity Matching},
  year =          {2025},
}

@article{cheng2025twinflow,
  author =        {Cheng, Zhenglin and Sun, Peng and Li, Jianguo and
                   Lin, Tao},
  journal =       {arXiv preprint arXiv:2512.05150},
  title =         {TwinFlow: Realizing One-step Generation on Large
                   Models with Self-adversarial Flows},
  year =          {2025},
}

@article{lin2025adversarial,
  author =        {Lin, Shanchuan and Yang, Ceyuan and Lin, Zhijie and
                   Chen, Hao and Fan, Haoqi},
  journal =       {arXiv preprint arXiv:2511.22475},
  title =         {Adversarial Flow Models},
  year =          {2025},
}

@inproceedings{zhu2025di,
  author =        {Zhu, Yuanzhi and Wang, Xi and
                   Lathuili{\`e}re, St{\'e}phane and Kalogeiton, Vicky},
  booktitle =     {Proceedings of the IEEE/CVF International Conference
                   on Computer Vision},
  pages =         {18606--18618},
  title =         {Di [M] o: Distilling masked diffusion models into
                   one-step generator},
  year =          {2025},
}

@inproceedings{zhu2024slimflow,
  author =        {Zhu, Yuanzhi and Liu, Xingchao and Liu, Qiang},
  booktitle =     {European Conference on Computer Vision},
  organization =  {Springer},
  pages =         {342--359},
  title =         {Slimflow: Training smaller one-step diffusion models
                   with rectified flow},
  year =          {2024},
}

@article{nguyen2025revisiting,
  author =        {Nguyen, Thanh and Yoo, Chang D},
  journal =       {arXiv preprint arXiv:2508.13904},
  title =         {Revisiting Diffusion Q-Learning: From Iterative
                   Denoising to One-Step Action Generation},
  year =          {2025},
}

@article{lv2025flow,
  author =        {Lv, Lei and Li, Yunfei and Luo, Yu and Sun, Fuchun and
                   Kong, Tao and Xu, Jiafeng and Ma, Xiao},
  journal =       {arXiv preprint arXiv:2506.12811},
  title =         {Flow-Based Policy for Online Reinforcement Learning},
  year =          {2025},
}

@article{chen2025one,
  author =        {Chen, Tianyi and Ma, Haitong and Li, Na and Wang, Kai and
                   Dai, Bo},
  journal =       {arXiv preprint arXiv:2507.23675},
  title =         {One-Step Flow Policy Mirror Descent},
  year =          {2025},
}

@inproceedings{li2025reinforcement,
  author =        {Qiyang Li and Zhiyuan Zhou and Sergey Levine},
  booktitle =     {The Thirty-ninth Annual Conference on Neural
                   Information Processing Systems},
  title =         {Reinforcement Learning with Action Chunking},
  year =          {2025},
  url =           {https://openreview.net/forum?id=XUks1Y96NR},
}

@article{schulman2017proximal,
  author =        {Schulman, John and Wolski, Filip and
                   Dhariwal, Prafulla and Radford, Alec and
                   Klimov, Oleg},
  journal =       {arXiv preprint arXiv:1707.06347},
  title =         {Proximal policy optimization algorithms},
  year =          {2017},
}

@article{brockman2016openaigym,
  author =        {Brockman, Greg and Cheung, Vicki and
                   Pettersson, Ludwig and Schneider, Jonas and
                   Schulman, John and Tang, Jie and Zaremba, Wojciech},
  journal =       {arXiv preprint arXiv:1606.01540},
  title =         {Openai gym},
  year =          {2016},
}

@article{towers2024gymnasium,
  author =        {Towers, Mark and Kwiatkowski, Ariel and Terry, Jordan and
                   Balis, John U and De Cola, Gianluca and
                   Deleu, Tristan and Goul{\~a}o, Manuel and
                   Kallinteris, Andreas and Krimmel, Markus and
                   KG, Arjun and others},
  journal =       {arXiv preprint arXiv:2407.17032},
  title =         {Gymnasium: A standard interface for reinforcement
                   learning environments},
  year =          {2024},
}

@inproceedings{todorov2012mujoco,
  author =        {Todorov, Emanuel and Erez, Tom and Tassa, Yuval},
  booktitle =     {2012 IEEE/RSJ international conference on intelligent
                   robots and systems},
  organization =  {IEEE},
  pages =         {5026--5033},
  title =         {Mujoco: A physics engine for model-based control},
  year =          {2012},
}

@article{zakka2025mujoco,
  author =        {Zakka, Kevin and Tabanpour, Baruch and Liao, Qiayuan and
                   Haiderbhai, Mustafa and Holt, Samuel and
                   Luo, Jing Yuan and Allshire, Arthur and Frey, Erik and
                   Sreenath, Koushil and Kahrs, Lueder A and others},
  journal =       {arXiv preprint arXiv:2502.08844},
  title =         {Mujoco playground},
  year =          {2025},
}

@article{tassa2018deepmind,
  author =        {Tassa, Yuval and Doron, Yotam and Muldal, Alistair and
                   Erez, Tom and Li, Yazhe and Casas, Diego de Las and
                   Budden, David and Abdolmaleki, Abbas and Merel, Josh and
                   Lefrancq, Andrew and others},
  journal =       {arXiv preprint arXiv:1801.00690},
  title =         {Deepmind control suite},
  year =          {2018},
}

@article{tunyasuvunakool2020dm_control,
  author =        {Tunyasuvunakool, Saran and Muldal, Alistair and
                   Doron, Yotam and Liu, Siqi and Bohez, Steven and
                   Merel, Josh and Erez, Tom and Lillicrap, Timothy and
                   Heess, Nicolas and Tassa, Yuval},
  journal =       {Software Impacts},
  pages =         {100022},
  publisher =     {Elsevier},
  title =         {dm\_control: Software and tasks for continuous
                   control},
  volume =        {6},
  year =          {2020},
}

@inproceedings{mahmood2019amass,
  author =        {Mahmood, Naureen and Ghorbani, Nima and
                   Troje, Nikolaus F and Pons-Moll, Gerard and
                   Black, Michael J},
  booktitle =     {Proceedings of the IEEE/CVF international conference
                   on computer vision},
  pages =         {5442--5451},
  title =         {AMASS: Archive of motion capture as surface shapes},
  year =          {2019},
}

@article{makoviychuk2021isaac,
  author =        {Makoviychuk, Viktor and Wawrzyniak, Lukasz and
                   Guo, Yunrong and Lu, Michelle and Storey, Kier and
                   Macklin, Miles and Hoeller, David and Rudin, Nikita and
                   Allshire, Arthur and Handa, Ankur and others},
  journal =       {arXiv preprint arXiv:2108.10470},
  title =         {Isaac gym: High performance gpu-based physics
                   simulation for robot learning},
  year =          {2021},
}

@inproceedings{luo2023perpetual,
  author =        {Luo, Zhengyi and Cao, Jinkun and Kitani, Kris and
                   Xu, Weipeng and others},
  booktitle =     {Proceedings of the IEEE/CVF International Conference
                   on Computer Vision},
  pages =         {10895--10904},
  title =         {Perpetual humanoid control for real-time simulated
                   avatars},
  year =          {2023},
}

@article{peng2018deepmimic,
  author =        {Peng, Xue Bin and Abbeel, Pieter and Levine, Sergey and
                   Van de Panne, Michiel},
  journal =       {ACM Transactions On Graphics (TOG)},
  number =        {4},
  pages =         {1--14},
  publisher =     {ACM New York, NY, USA},
  title =         {Deepmimic: Example-guided deep reinforcement learning
                   of physics-based character skills},
  volume =        {37},
  year =          {2018},
}

@inproceedings{fan2025online,
  author =        {Fan, Jiajun and Shen, Shuaike and Cheng, Chaoran and
                   Chen, Yuxin and Liang, Chumeng and Liu, Ge},
  booktitle =     {The Thirteenth International Conference on Learning
                   Representations},
  title =         {Online reward-weighted fine-tuning of flow matching
                   with wasserstein regularization},
  year =          {2025},
}

\newpage
\appendix
\renewcommand{\thetable}{\thesection.\arabic{table}}
\onecolumn
\section{Code and Supplementary Videos}
\label{appendix:code_and_videos}

The source code for ReFPO is included in the supplementary materials to ensure the reproducibility of all experimental results. 

Additionally, we provide supplementary videos demonstrating the performance of ReFPO and FPO across MuJoCo Playground and Humanoid Control tasks. These videos showcase the high-fidelity trajectories achieved during one-step inference ($N=1$) and provide direct comparisons with the baseline methods.

\section{Training Settings}
\label{appendix:training_settings}

\subsection{GridWorld}
\label{appendix:gridworld}

All GridWorld experiments were conducted using an \textbf{NVIDIA GeForce RTX 4090 GPU}. To ensure a rigorous and fair comparison, we maintained identical hyperparameters for both the vanilla FPO baseline and our proposed ReFPO method with 5 seeds, the only exception being the Reflow regularization coefficient $\lambda$ utilized in ReFPO.Each run required approximately 0.5 GPU hours, resulting in roughly 5 RTX 4090 GPU hours in total for the GridWorld experiments. A comprehensive summary of these hyperparameters is provided in Table~\ref{tab:gridworld_hyperparams}.
\begin{table}[ht]
\caption{Hyperparameters for GridWorld Experiments}
\label{tab:gridworld_hyperparams}
\vskip 0.15in
\begin{center}
\begin{small}
\begin{sc}
\begin{tabular}{lc}
\toprule
Hyperparameter & Value \\
\midrule
Optimizer & Adam \\
Learning Rate & $3 \times 10^{-4}$ \\
Total Timesteps & 260,000 \\
Timesteps per Batch & 2048 \\
Minibatch Size & 341 (approx. $2048/6$) \\
Updates per Iteration & 10 \\
Max Episode Length & 200 \\
Discount Factor ($\gamma$) & 0.99 \\
GAE Parameter ($\lambda_{GAE}$) & 0.98 \\
PPO Clip Range ($\epsilon$) & 0.2 \\
Max Grad Norm & 0.5 \\
MC Samples for CFM ($N_{mc}$) & 50 \\
Reflow Coefficient & 0.1 (0.0 for FPO) \\
ODE Solver & Euler \\
Inference Steps ($N$) & 1, 10, 20 \\
\bottomrule
\end{tabular}
\end{sc}
\end{small}
\end{center}
\vskip -0.1in
\end{table}

\subsection{MuJoCo Playground}
\label{appendix:mujoco}

For the MuJoCo Playground benchmarks, we evaluated our methods across 10 continuous control tasks using an \textbf{NVIDIA GeForce RTX 4090 GPU}. All models were trained for a total of 100 million environment steps to observe asymptotic performance and flow stability. For PPO, we utilized a standard Gaussian policy as a baseline, training it for the same 100 million environment steps with 5 seeds to align the experimental configuration across all methods.Each task-seed-configuration run required approximately 0.08 GPU hours on average. Therefore, the total compute for the MuJoCo Playground experiments was approximately 44 RTX 4090 GPU hours.

Similar to the GridWorld experiments, the hyperparameters for FPO and ReFPO are identical except for the Reflow regularization coefficient $\lambda$ and Output mode. Regarding the Output mode, it is worth noting that while the theoretical formulations of both FPO and ReFPO are inherently based on velocity prediction, the vanilla FPO baseline adopts a modified output mode specifically tuned to maximize cumulative rewards in certain environments. However, our empirical analysis indicated that such a modification is suboptimal for the ReFPO framework and compromises the stability of the learned flows. Therefore, we strictly adhere to the standard velocity prediction in our method to maintain consistency with the underlying flow-matching objective, ensuring a more stable and theoretically grounded training process.
\begin{table}[!htbp]
    \centering
    \small
    \caption{\textbf{Hyperparameters for PPO, FPO/ReFPO, Flow-GRPO, and ReinFlow.}}
    \label{tab:mujoco_hyperparameters}

    % ================= Row 1 =================
    % --- (a) PPO ---
    \begin{minipage}[t]{0.48\textwidth}
        \centering
        \caption*{(a) PPO}
        \begin{tabular}{ll}
            \toprule
            \textbf{Hyperparameter} & \textbf{Value} \\
            \midrule
            Discount factor $(\gamma)$     & 0.995 \\
            GAE $\lambda$                  & 0.95 \\
            Value loss coefficient         & 0.25 \\
            Entropy coefficient            & 0.01 \\
            Reward scaling                 & 10.0 \\
            Normalize advantage            & True \\
            Normalize observations         & True \\
            Action repeat                  & 1 \\
            Unroll length                  & 30 \\
            Batch size                     & 1024 \\
            Number of minibatches          & 32 \\
            Number of updates per batch    & 16 \\
            Number of timesteps            & 100M \\
            Policy network                 & MLP (4, 32) \\
            Value network                  & MLP (5, 256) \\
            Optimizer                      & Adam \\
            \bottomrule
        \end{tabular}
    \end{minipage}
    \hfill
    % --- (b) FPO / ReFPO ---
    \begin{minipage}[t]{0.48\textwidth}
        \centering
        \caption*{(b) FPO / ReFPO}
        \begin{tabular}{ll}
            \toprule
            \textbf{Hyperparameter} & \textbf{Value} \\
            \midrule
            Discount factor $(\gamma)$     & 0.995 \\
            GAE $\lambda$                  & 0.95 \\
            Value loss coefficient         & 0.25 \\
            Clipping $\epsilon$            & 0.05 \\
            Reward scaling                 & 10.0 \\
            Reflow coefficient             & 0.04 (ReFPO) \\
            Normalize advantage            & True \\
            Normalize observations         & True \\
            Flow steps $(T)$               & 10 \\
            Samples per action             & 8 \\
            Output mode                    & velocity/eps \\
            Time embed dim                 & 8 \\
            Unroll length                  & 30 \\
            Batch size                     & 1024 \\
            Number of minibatches          & 32 \\
            Number of updates per batch    & 16 \\
            Number of timesteps            & 100M \\
            Policy network                 & MLP (4, 32) \\
            Value network                  & MLP (5, 256) \\
            Optimizer                      & Adam \\
            \bottomrule
        \end{tabular}
    \end{minipage}

    \vspace{1.0em}

    % ================= Row 2 =================
    % --- (c) GRPO ---
    \begin{minipage}[t]{0.48\textwidth}
        \centering
        \caption*{(c) Flow-GRPO}
        \begin{tabular}{ll}
            \toprule
            \textbf{Hyperparameter} & \textbf{Value} \\
            \midrule
            Discount factor $(\gamma)$     & 0.995 \\
            Reward scaling                 & 10.0 \\
            Normalize observations         & True \\
            Flow steps $(T)$               & 10 \\
            Output mode                    & $u$ \\
            Time embed dim                 & 8 \\
            Loss mode                      & denoising MDP \\
            Group size                     & 8 \\
            Advantage clipping max         & 8.0 \\
            Clipping $\epsilon$            & 0.1 \\
            KL coefficient $\beta$         & $5 \times 10^{-3}$ \\
            Noise level                    & 1.0 \\
            Feather std                    & 0.05 \\
            Policy output scale            & 0.5 \\
            Unroll length                  & 30 \\
            Number of environments         & 2048 \\
            Batch size                     & 1024 \\
            Number of minibatches          & 32 \\
            Number of updates per batch    & 8 \\
            Number of timesteps            & 200M \\
            Learning rate                  & $1 \times 10^{-4}$ \\
            Max gradient norm              & 1.0 \\
            Policy network                 & MLP (4, 32) \\
            Value network                  & N/A \\
            Optimizer                      & Adam \\
            \bottomrule
        \end{tabular}
    \end{minipage}
    \hfill
    % --- (d) ReinFlow ---
    \begin{minipage}[t]{0.48\textwidth}
        \centering
        \caption*{(d) ReinFlow}
        \begin{tabular}{ll}
            \toprule
            \textbf{Hyperparameter} & \textbf{Value} \\
            \midrule
            Discount factor $(\gamma)$     & 0.99 \\
            GAE $\lambda$                  & 0.95 \\
            Learning rate                  & $3 \times 10^{-5}$ \\
            Optimizer                      & Adam \\
            Batch size                     & 1024 \\
            Number of environments         & 2048 \\
            Unroll length                  & 30 \\
            Number of minibatches          & 32 \\
            Number of updates per batch    & 8 \\
            Number of timesteps            & 200M \\
            Reward scaling                 & 3.0 \\
            Normalize observations         & True \\
            Normalize advantage            & True \\
            Value loss coefficient         & 0.6 \\
            Entropy coefficient            & 0.05 \\
            PPO clipping $\epsilon$        & 0.10 \\
            \midrule
            Horizon steps                  & 4 \\
            Inference denoising steps      & 4 \\
            Fine-tuning denoising steps    & 2 \\
            Timestep embedding dim         & 16 \\
            Action range                   & $[-1, 1]$ \\
            Noise scheduler                & Linear \\
            Sampling denoising std min     & 0.10 \\
            Log-prob denoising std range   & $[0.08, 0.16]$ \\
            \bottomrule
        \end{tabular}
    \end{minipage}
\end{table}
For ReFPO, we applied a constant regularization coefficient of $\lambda = 0.04$, which was empirically found to provide the best balance between path straightness and policy expressivity in these high-dimensional tasks. The detailed hyperparameter configurations are summarized in Table~\ref{tab:mujoco_hyperparameters}, where we also include the GRPO and ReinFlow settings.

% For FPO baseline, we report the stronger engineering-tuned output mode used in our implementation, which empirically improves its cumulative reward over the standard velocity-prediction version. ReFPO, in contrast, uses the standard velocity-prediction parameterization because it is directly aligned with the flow-matching objective and yields more stable flow geometry. Thus, this design choice favors the FPO baseline rather than ReFPO, making the comparison conservative with respect to our proposed regularizer.

Similar to the GridWorld experiments, the hyperparameters for FPO and ReFPO are identical except for the Reflow regularization coefficient $\lambda$ and the output parameterization. Since output parameterization can affect flow-policy performance, we explicitly ablate this choice for the FPO baseline in Table~\ref{tab:fpo_output_mode_ablation}. In our implementation, the noise-prediction variant used for the main FPO baseline is stronger than the standard velocity-prediction variant, achieving higher 10-step and 1-step rewards and slightly better flow diagnostics.

ReFPO uses the canonical velocity-prediction parameterization because the proposed Reflow regularizer is formulated directly on the learned velocity field. Therefore, the output-mode choice does not give ReFPO an advantage: the main comparison gives FPO its empirically stronger output mode, while evaluating ReFPO in the formulation aligned with its regularizer. This makes the comparison conservative with respect to the proposed Reflow regularization, rather than explaining away the observed gains.

\begin{table}[ht]
  \caption{Output parameterization ablation for FPO on MuJoCo Playground.}
  \label{tab:fpo_output_mode_ablation}
  \centering
  \begin{small}
  \begin{tabular}{lcccc}
    \toprule
    \textbf{Method} & \textbf{10-step Reward} & \textbf{1-step Reward} & \textbf{Straightness} & \textbf{Explosion} \\
    \midrule
    FPO & $641 \pm 140$ & $565 \pm 160$ & $0.0475$ & $0.00365$ \\
    FPO-velocity & $602 \pm 168$ & $533 \pm 172$ & $0.0515$ & $0.00389$ \\
    \bottomrule
  \end{tabular}
  \end{small}
\end{table}
\subsection{Humanoid Control}

All humanoid control experiments were conducted using an \textbf{NVIDIA A100 GPU}, with each individual training run requiring approximately \textbf{14 hours} to complete. To ensure a rigorous and fair comparison, we maintained identical hyperparameters for both the vanilla FPO baseline and our proposed ReFPO method, with the only exception being the Reflow regularization coefficient $\lambda$ associated with the Reflow loss, which was utilized exclusively in ReFPO. A comprehensive summary of these hyperparameters is provided in Table~\ref{tab:humanoid_hyperparams}.

\begin{table}[ht]
\centering
\caption{Hyperparameters for humanoid control.}
\label{tab:humanoid_hyperparams}
\begin{tabular}{ll|ll}
\hline
\textbf{Hyperparameter} & \textbf{Value} & \textbf{Hyperparameter} & \textbf{Value} \\
\hline
\multicolumn{4}{c}{\textit{Policy Settings}} \\
\hline
Hidden size & 512 & Solver step size & 0.1 \\
Action perturbation std & 0.05 & Target KL divergence & None \\
Number of environments & 4096 & Normalize advantage & True \\
Reflow coefficient $\lambda$ & \multicolumn{3}{l}{0.1 for ReFPO; 0.0 for FPO} \\
\hline
\multicolumn{4}{c}{\textit{Training Settings}} \\
\hline
Batch size & 131072 & Minibatch size & 32768 \\
Learning rate & 0.0001 & LR annealing & False \\
LR decay rate & 1.5e-4 & LR decay floor & 0.2 \\
Update epochs & 4 & L2 regularization coef. & 0.0 \\
GAE lambda & 0.2 & Discount factor ($\gamma$) & 0.98 \\
Clipping coefficient & 0.01 & Value function coefficient & 1.2 \\
Clip value loss & True & Value loss clip coefficient & 0.2 \\
Max gradient norm & 10.0 & Entropy coefficient & 0.0 \\
Discriminator coefficient & 5.0 & Bound coefficient & 10.0 \\
\hline
\end{tabular}
\end{table}

\section{Additional Stability Diagnostics}
\label{appendix:jacobian_ratio_diagnostics}

We include Jacobian and policy-ratio diagnostics as additional stability measures for the MuJoCo Playground experiments. These quantities are not used as primary evaluation metrics, but they provide a complementary view of the local behavior of the learned flow and the magnitude of policy-ratio fluctuations. As shown in Table~\ref{tab:jacobian_policy_ratio_diagnostics}, ReFPO exhibits a lower Jacobian explosion rate, a lower policy-ratio explosion rate, and smaller Jacobian norms than FPO.

\begin{table}[!htbp]
    \centering
    \small
    \caption{Jacobian and policy-ratio diagnostic statistics on MuJoCo Playground.
    }
    \label{tab:jacobian_policy_ratio_diagnostics}
    \begin{tabular}{lcccc}
        \toprule
        \textbf{Method} &
        \textbf{Jac. Expl.} &
        \textbf{Ratio Expl.} &
        \textbf{Mean Jac. Norm} &
        \textbf{Jac. Norm P95} \\
        \midrule
        ReFPO & 0.005654 & 0.00189 & 0.4881 & 0.9067 \\
        FPO   & 0.008724 & 0.00365 & 0.5551 & 1.219  \\
        \bottomrule
    \end{tabular}
\end{table}

\clearpage
\section{Algorithmic details}
\label{app:algorithm}
\setcounter{algorithm}{0}
\renewcommand{\thealgorithm}{\thesection.\arabic{algorithm}}
\begin{algorithm}[H]
   \caption{Reflow-regularized Flow Matching Policy Gradients (ReFPO)}
   \label{alg:refpo}
\begin{algorithmic}
   \STATE {\bfseries Require:} Policy parameters $\theta$, value parameters $\phi$, clip parameter $\epsilon$, MC samples $N_{\text{mc}}$, Reflow coefficient $\lambda$
   \WHILE{not converged}
      \STATE Collect trajectories using current flow policy and compute advantages $\hat{A}_t$
      \STATE For each action, store $N_{\text{mc}}$ timestep-noise pairs $\{(\tau_i, \epsilon_i)\}$ and compute $\ell_{\theta}(\tau_i, \epsilon_i)$
      \STATE $\theta_{\text{old}} \leftarrow \theta$
      \FOR{each optimization epoch}
         \STATE Sample mini-batch from collected trajectories
         \FOR{each state-action pair $(o_t, a_t)$ and corresponding MC samples $\{(\tau_i, \epsilon_i)\}$}
            \STATE Compute flow matching loss $\ell_{\theta}(\tau_i, \epsilon_i)$ using stored $(\tau_i, \epsilon_i)$
            \STATE $\hat{r}_{\theta} \leftarrow \exp \left( -\frac{1}{N_{\text{mc}}} \sum_{i=1}^{N_{\text{mc}}} (\ell_{\theta}(\tau_i, \epsilon_i) - \ell_{\theta_{\text{old}}}(\tau_i, \epsilon_i)) \right)$
            \STATE $\mathcal{L}^{\text{Reflow}}(\theta) \leftarrow \frac{1}{N_{\text{mc}}} \sum_{i=1}^{N_{\text{mc}}} \ell_{\theta}(\tau_i, \epsilon_i)$ \COMMENT{Reuse $\ell_{\theta}$ from above}
            \STATE $L^{\text{FPO}}(\theta) \leftarrow -\min(\hat{r}_{\theta}\hat{A}_t, \text{clip}(\hat{r}_{\theta}, 1 \pm \epsilon)\hat{A}_t)$
            \STATE $L^{\text{ReFPO}}(\theta) \leftarrow L^{\text{FPO}}(\theta) + \lambda \mathcal{L}^{\text{Reflow}}(\theta)$
         \ENDFOR
         \STATE $\theta \leftarrow \text{Optimizer}(\theta, \nabla_{\theta} \sum L^{\text{ReFPO}}(\theta))$
      \ENDFOR
      \STATE Update value function parameters $\phi$ like standard PPO
   \ENDWHILE
\end{algorithmic}
\end{algorithm}

\end{document}